\documentclass[10pt,twocolumn,letterpaper]{article}

\usepackage[pagenumbers]{cvpr} 

\usepackage{graphicx}
\usepackage{amsmath}
\usepackage{amssymb}
\usepackage{booktabs, comment}

\usepackage{url}
\usepackage[utf8]{inputenc} 
\usepackage[T1]{fontenc}    
\usepackage[table,xcdraw,dvipsnames]{xcolor}
\usepackage{subcaption}
\usepackage[normalem]{ulem} 
\usepackage[export]{adjustbox}
\usepackage{amsfonts}       
\usepackage{nicefrac}       
\usepackage{microtype}      
\usepackage{gensymb}
\usepackage{tabularx}
\usepackage{siunitx}
\usepackage{enumitem}
\usepackage{lineno}
\usepackage{pifont}
\usepackage{upgreek}
\usepackage{bbm}
\usepackage{scalerel}

\usepackage[dvipsnames]{xcolor}

\usepackage[font=small,labelfont=bf]{caption}

\usepackage{stfloats}

\newcommand{\RN}[1]{%
  \textup{\uppercase\expandafter{\romannumeral#1}}%
}

\usepackage[dvipsnames]{xcolor}

\newcommand{\KGnote}[1]{{\color{blue}{\bf KG: }#1}}

\definecolor{navy}{rgb}{0.7, 0.1, 0.7}
\definecolor{burgundy}{RGB}{144,0,32}
\definecolor{green2}{RGB}{0,160,0}

\newcommand{\ZAnote}[1]{{\color{teal}{\bf ZA: }#1}} 

\newcommand{\SMnote}[1]{{\color{burgundy}{\bf SM: }#1}} 

\renewcommand{\KGnote}[1]{{\color{magenta}}} 
\renewcommand{\ZAnote}[1]{{\color{teal}}} 
\renewcommand{\SMnote}[1]{{\color{red}}} 

\definecolor{cvprblue}{rgb}{0.21,0.49,0.74}
\usepackage[pagebackref,breaklinks,colorlinks,citecolor=cvprblue]{hyperref}

\usepackage[capitalize]{cleveref}
\crefname{section}{Sec.}{Secs.}
\Crefname{section}{Section}{Sections}
\Crefname{table}{Table}{Tables}
\crefname{table}{Tab.}{Tabs.}

\def\paperID{xxxxx} 
\def\confName{CVPR}
\def\confYear{2024}

\usepackage{fancyhdr}

\fancypagestyle{firstpage}{%
  \fancyhf{}
  \chead{\small In Proceedings of IEEE/CVF Conference on Computer Vision and Pattern Recognition (CVPR), 2024}
}

\title{Learning Spatial Features from Audio-Visual Correspondence\\in Egocentric Videos}

\author{%
Sagnik Majumder$^{1,2}$ \hspace{3mm} \hspace{3mm}  Ziad Al-Halah$^{3}$ \hspace{3mm} \hspace{3mm} Kristen Grauman$^{1,2}$\\
$^1$UT Austin \hspace{3mm} $^2$FAIR, Meta \hspace{3mm} $^3$University of Utah 
}

\begin{document}
\maketitle

\thispagestyle{firstpage}

\begin{abstract}
We propose a self-supervised method for learning representations based on spatial audio-visual correspondences in egocentric videos. 
{Our} method 
{uses} a masked auto-encoding framework to synthesize masked binaural {(multi-channel)} audio through the synergy of audio and vision, thereby learning useful spatial relationships between the two modalities. We use our pretrained features to tackle two downstream video tasks requiring spatial understanding in social scenarios: active speaker detection and spatial audio denoising.  
Through extensive experiments, we show that our features are generic enough to improve over multiple state-of-the-art baselines on {both tasks on} two challenging egocentric video datasets {that offer binaural audio},  EgoCom and EasyCom. Project: 
\url{http://vision.cs.utexas.edu/projects/ego_av_corr}.
\end{abstract}
\vspace{-0.25cm}

\section{Introduction}\label{sec:intro}
Egocentric videos provide a first-person view of how we perceive and interact with our surroundings in daily life, 
and they are pushing a new frontier in multi-modal learning~\citep{Damen2022RESCALING, journals/corr/abs-1804-09626, grauman2022ego4d, huh2023epic}.  A key aspect of ego-video is that it can provide a rich stream of first-person spatial 
audio\footnote{{Throughout we use the term \emph{spatial audio} to refer to 
binaural audio, including the special case of two-channel \emph{binaural} audio as perceived by two human ears.  In contrast, single-channel \emph{monaural} audio lacks spatial information.}}  alongside 
visual frames when the audio is captured with multiple microphones~\citep{9200754, donley2021easycom}. The coupling of such audio and vision provides strong spatial information about the sound sources (\eg where the sound sources are, if they are in motion or not) in the context of the surrounding physical space (\eg how big or small the room is, if there is a large wall nearby), as well as the camera wearer's attention in the scene 
{revealed by} how they move their head. 

\begin{figure}[t] 
    \centering
    \includegraphics[width=1\linewidth]{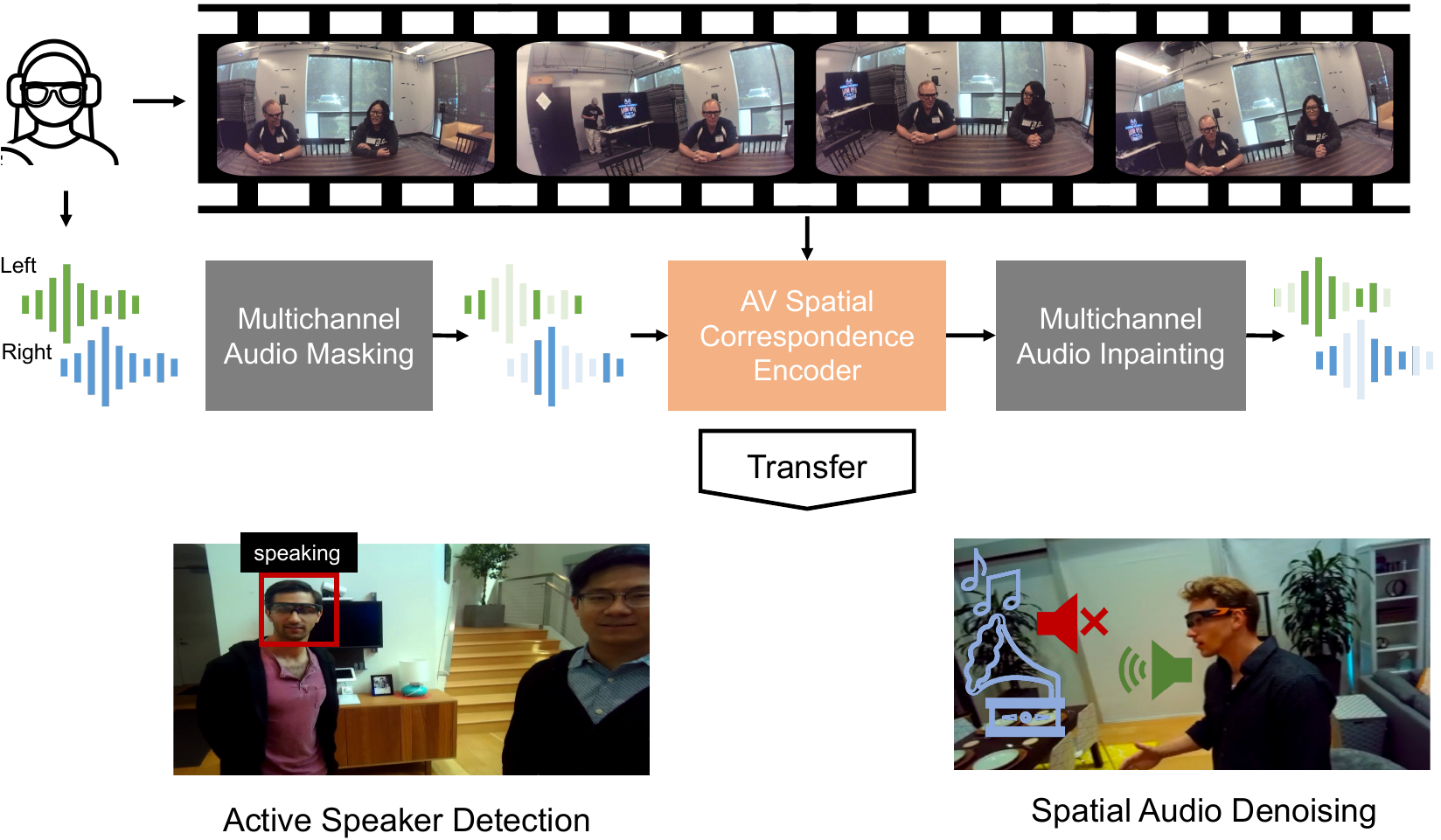}
    \caption{
Given an egocentric video and binaural audio, we aim to learn spatial correspondences between vision and audio by solving the pretext task of inpainting segments of the binaural audio.
The features benefit  downstream {social} tasks {where spatial localization is important:} active speaker detection and audio denoising.
    }
\label{fig:intro}
    \vspace{-0.25cm}
\end{figure}

Such spatial cues are especially important for social settings of multiple people talking to each other, where it is valuable to be able to {1)} focus on the voice(s) of interest from among various competing sounds and {2)} understand where people are directing their
speech activity, for better comprehension and communication. In this way, future AR applications in conversational settings could allow a hearing-impaired person to determine who is speaking in order to redirect their attention, or enhance the received audio to make it more intelligible for any listener.

We argue that this creates the need for human-centric spatially-grounded understanding of audio-visual events. {Can we} 
learn representations from video that capture audio-visual events \emph{in the context of the persistent physical space of the environment and the human speakers in it}? Such representations {would be}
useful for answering questions like ``who is speaking right now?" and ``what would the voices sound like without 
the audio noise (distractor sounds)?" 
While the former requires inferring the source location for a voice in the scene, the latter requires understanding how the perceived audio is a function of the source locations, the listener, and the surrounding environment.

Despite the significance of these problems, {today's models for audio-visual learning are not human-centric and they lack spatial grounding.}
{On the one hand,} current audio-visual representation learning methods exclusively tackle exocentric  (third-person) video {with monaural audio} ~\citep{Owens2016AmbientSP, owens2018audio, korbar2018cooperative, afouras2020self,georgescu2022audiovisual, huang2022mavil, gong2023contrastive}.
{That domain} 
sidesteps challenges inherent to ego-video arising from the camera wearer's head motion and relatively limited field of view.  
{On the other hand,} the limited prior work exploring self-supervised objectives using multi-channel audio and video~\cite{gao20192,yang2020telling,morgado2020learning, gao2020visualechoes} are also outside of the egocentric and social contexts {(e.g., for sounds in empty homes~\cite{gao2020visualechoes}, musical instruments~\cite{gao20192}, or single human speakers~\cite{yang2020telling}), where the need for spatial understanding of multiple sound sources is limited}.

We propose to learn audio-visual representations
via spatial correspondence between an egocentric video and its
binaural audio, for analyzing social (conversational) settings. In particular, we design a novel pretext task where the goal is to inpaint binaural (two-channel) audio using both video and audio. 
Given 
a social egocentric video clip with binaural audio, we mask segments of it and train a model based on 
masked autoencoding (MAE)~\citep{feichtenhofer2022masked,he2022masked, tong2022videomae, huang2022masked, Baade2022MAEASTMA} 
to predict the missing segments on the basis of the video and the unmasked segments in the audio. See Figure~\ref{fig:intro} (top). 
Additionally, we introduce a {novel spatial audio masking strategy that facilitates learning strong audio-visual spatial correspondences while maintaining learning stability when vision alone is insufficient for the binauralization task.}
Once trained, our model's encoder provides 
spatial audio-visual 
features that can be used to address multiple downstream tasks, {as we demonstrate} using multiple {different} backbones and social egocentric video datasets.

In particular, motivated by the AR  applications discussed above, we validate our feature learning method on two downstream social egocentric tasks that require 
strong audio-visual spatial reasoning: 
\textbf{1) active speaker detection:} predicting which person in the field of view of an egocentric video is speaking, and \textbf{2) spatial audio denoising:} separating audio noise (any sounds from non-speakers) from the input audio. 
See Figure~\ref{fig:intro} (bottom).  We test the generality of our method by evaluating on two social egocentric video datasets, EgoCom~\citep{9200754} and EasyCom~\citep{donley2021easycom}{---to our knowledge, the only two publicly available video datasets with binaural sound and social settings.} On both, our method significantly outperforms multiple 
state-of-the-art task-specific and audio-visual spatial feature learning models.

\section{Related Work}

\paragraph{Audio-visual self-supervised pretraining}
Past work~\citep{Ngiam2011MultimodalDL, Owens2016AmbientSP, Arandjelovi2017LookLA, korbar2018cooperative, owens2018audio, owens2016visually, afouras2020self, morgado2020learning} extensively studies the synergy of vision and audio for learning representations through self-supervision. 
They explore using both modalities to construct pretext tasks based on synthesis~\cite{Owens2016AmbientSP,owens2016visually}, alignment~\cite{Arandjelovi2017LookLA, korbar2018cooperative, owens2018audio,afouras2020self,gao2020visualechoes}, and masked auto-encoding (MAE)~\cite{gong2023contrastive, georgescu2022audiovisual, huang2022mavil}, and they focus on \emph{semantic} downstream tasks like audio-visual event classification and retrieval.
However, none of these methods are designed to extract spatial cues from video and multi-channel audio, nor do they analyze the social egocentric setting. On the contrary, we propose 
self-supervised learning of spatial audio-visual features from egocentric video. 
Further, different from existing MAE-style models~\citep{gong2023contrastive, georgescu2022audiovisual, huang2022mavil}, we propose a specialized masking strategy that better learns spatial audio-visual {cues}. 
{Our masking idea promotes the encoding of spatial and semantic information in the learned multimodal representation, thereby improving its effectiveness for transfer learning in downstream tasks that require nuanced reasoning about both \textit{what} and \textit{where} aspects, such as active speaker detection and spatial audio-visual denoising. This differs from previous methods~\cite{georgescu2022audiovisual, huang2022mavil, gong2023contrastive}, which mainly use a learning objective that emphasizes the encoding of semantic cues and tailor to tasks like multimodal event classification or retrieval.}

\vspace{-0.3cm}
\paragraph{Audio-visual spatial correspondence learning}
Learning the \emph{spatial} alignment between video and audio is important for self-supervision~\citep{yang2020telling, morgado2020learning, 9790080, sato2022self-supervised}, 
audio generation~\citep{NEURIPS2018_01161aaa, gao20192, Rachavarapu_2021_ICCV, chen2022visual, majumder2022fewshot,morrone-audio-inpainting,zhou-audio-inpainting},
audio-visual embodied learning~\citep{chen2020soundspaces, majumder2021move2hear, majumder2022active, chen2022soundspaces} and 3D scene mapping~\citep{purushwalkam2021audio, majumder2023chat2map}. However, these methods are either restricted to exocentric settings~\citep{NEURIPS2018_01161aaa, yang2020telling, morgado2020learning, gao20192, Rachavarapu_2021_ICCV, sato2022self-supervised, chen2022visual}, or else tackle egocentric settings~\citep{ majumder2022active, majumder2022fewshot, chen2022soundspaces, majumder2023chat2map} in simulated 3D environments that lack realism and diversity, both in terms of the audio-visual content of the videos {(no people are visible, no objects are moving)} and {their lack of} continuous camera motion 
from a camera-wearer's physical movements. 
In contrast, we learn an audio-visual representation
from real-world egocentric video.

More closely related to our work are Telling Left from Right (TLR)~\citep{yang2020telling}, 
2.5D Visual Sounds (2.5D-VS)~\citep{gao20192}, and audio-visual stereo sound ranking (SSR)~\citep{sato2022self-supervised}, 
all of which learn spatial audio-visual features, 
albeit for exocentric data only. 
TLR predicts whether the left and right binaural channels are swapped, and SSR ranks the similarity of video to different stereo audio samples through self-supervision---both of which 
provide only coarse spatial information about the scene.
2.5D-VS learns to ``lift" the mono input to binaural audio, which can be underconstrained from the single-channel audio and video alone.
We design a novel pretext task using audio-visual inpainting of 
binaural audio, which is both fine-grained (requiring to capture subtleties about the arrangement of speakers in the environment) and, through our novel masking strategy, 
exposes better multi-modal constraints that improve {learning stability and} performance.
Our results in Sec.~\ref{sec:experiments} show our model's advantages over 
all three prior methods~\cite{gao20192,yang2020telling, sato2022self-supervised}.

\vspace{-0.3cm}
\paragraph{Active speaker detection }
Active speaker detection (ASD) entails predicting the active speaker(s) from among all detected faces in a video, and 
{is}
a special case of generic 2D sound localization~\citep{NIPS1999_b618c321, 
owens2018audio,
yang2022camera, jiang2022egocentric, 
mo2022localizing, 
gan2019self}.
While early ASD methods rely on lip 
{motion}
and facial gestures~\citep{Everingham2006HelloMN}, 
recent methods employ ensemble networks~\citep{Alcazar2020ActiveSI} or 3D CNNs~\citep{Sharma2020CrossmodalLF, tao2021someone, kopuklu2021design},
relation context modules~\citep{zhang2021unicon}, attention ~\citep{Alcazar2020ActiveSI, tao2021someone}, or graph neural networks~\citep{LeonAlcazar2021MAASMA, min2022learning}.
Multi-channel audio improves ASD~\citep{jiang2022egocentric},
but requires privileged information of the speaker's pose for training.
{Recent work explores using supervised learning to infer not only who is talking, but also to whom a camera-wearer is listening~\cite{ryan2023egocentric}.}
{In contrast,}
our goal is to learn spatial audio-visual features purely from in-the-wild egocentric {video}
through self-supervision---features generic enough to benefit multiple ASD models, as we
{show}
for both TalkNet~\citep{tao2021someone} and SPELL~\cite{min2022learning}.

\vspace{-0.3cm}
\paragraph{Spatial audio denoising}
Audio denoising, which requires separating a target sound from noise, has traditionally been studied with single-channel (non-spatial) audio, both in the audio-only setting~\citep{10.1007/978-3-540-74494-8_52, Spiertz09source-filterbased, Virtanen07monauralsound, 6853860} and audio-visual settings~\citep{NIPS2000_11f524c3, pu2017audio, 7951787,  afouras2018conversation, ephrat2018looking, owens2018audio, gao2021visualvoice, gao2019co,  gao2018learning}. 
Using spatial audio captured with multiple microphones~\citep{nakadai2002real, Yilmaz04blindseparation, duong2010under} 
naturally makes the task simpler.
Different from the above, we learn task-agnostic audio-visual spatial features. That is, our contribution is the feature learning idea (which benefits both denoising and ASD), rather than a novel denoising approach.

\section{Learning spatial features from egocentric audio-visual correspondence}
\label{sec:task}

\begin{figure*}[!th] 
    \centering
    \includegraphics[width=0.80\linewidth]{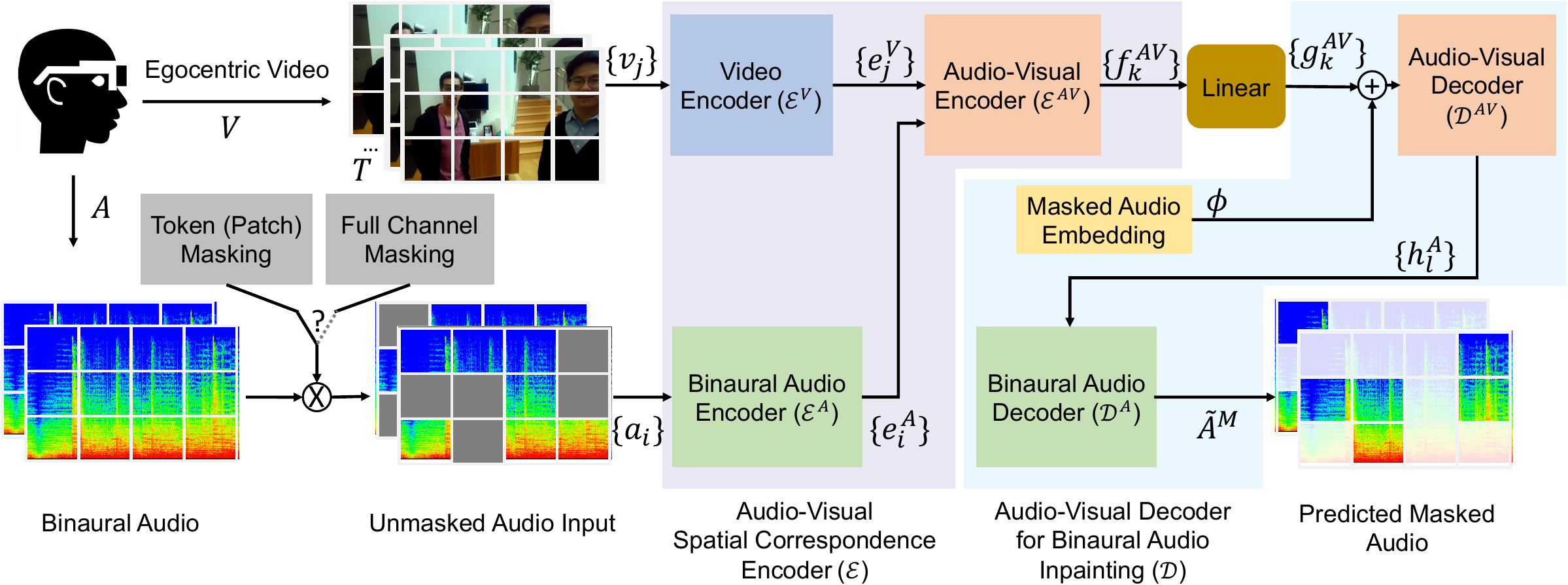}
    \caption{Our model learns the spatial correspondence between vision and binaural audio by inpainting masked tokens of the audio channels through the use of an audio-visual encoder-decoder model. {We combine random token masking (which requires solving a more local binauralization task) with complete audio channel masking (which requires more global cues to synthesize unseen binaural segments).} 
    For downstream evaluation, we fuse the features from the audio-visual encoder with the backbones for downstream tasks, and finetune them.
}
\vspace*{-0.15in}
\label{fig:model}
\end{figure*}

The spatial sound perceived in an egocentric 
{setup}
is shaped by the environment in which it is emitted and 
{its}
source location relative to the camera-wearer.
Based on this knowledge, we 
hypothesize
that trying to solve the pretext task of audio-visual inpainting of binaural audio---{that is,} 
{synthesizing}
missing audio segments
by extracting {related visual} 
{cues}
about the scene and the source location---can lead to learning useful audio-visual spatial correspondences. To validate our hypothesis, we propose a novel feature-learning task.

Formally, we consider an egocentric video clip $C = \big(V, A\big)$, where $V$ and $A$ refer to the visual and binaural audio streams, respectively. See Fig.~\ref{fig:model} left. The visual clip $V$ comprises $T$ frames, such that $V = \big\{V_{1}, \ldots,  V_{T}\big\}$.
We generate a set of visual tokens $\hat{V}$ by splitting $V$ into $P$ 
non-overlapping tubelets, 
such that $\hat{V} = \big\{\hat{V}_1, \ldots, \hat{V}_P\big\}$, where $\hat{V}_k$ denotes the $k^{th}$ tubelet consisting of a contiguous sequence of 
patches spanning all $T$ frames.
We represent the binaural audio $A$ as {two} Mel-spectrograms~\citep{huang2022masked}, $A = \big\{A^L, A^R\big\}$, where $A^L$ and $A^R$ are the spectrograms for the left and right channels, respectively. 
We create a set of audio tokens $\hat{A}$ by splitting $A$ into $Q$ non-overlapping patches,
such that $\hat{A} = \big\{\hat{A}_1, \ldots, \hat{A}_Q \big\}$. 

Next, we mask a portion of the audio tokens in $\hat{A}$ and obtain complementary subsets of masked and unmasked tokens, $\hat{A}^M$ and $\hat{A}^U$, respectively, where $\hat{A}^M = \big\{\ddot{A}_1, \ldots, \ddot{A}_S\big\}$, $\hat{A}^U = \big\{\bar{A}_1, \ldots, \bar{A}_{Q-S}\big\}$, and $S$ is the number of masked tokens. Given $\big\{\hat{V},  \hat{A}^M, \hat{A}^U\big\}$, we aim to learn a self-supervised model $\mathcal{F}$ comprising an encoder $\mathcal{E}$ and decoder $\mathcal{D}$, such that $\mathcal{F} = \mathcal{D} \circ \mathcal{E}$ and $\mathcal{F}(\hat{V}, \hat{A}^U) = \tilde{A}^M$, where $\tilde{A}^M$ is an estimate of the masked audio tokens in $\hat{A}^M$. By training on this pretext task, our encoder $\mathcal{E}$ can learn rich audio-visual spatial correspondences that can be leveraged for multiple downstream tasks that require the synergy of vision and spatial audio, as we show in results.

In our method (see Fig.~\ref{fig:model}), $\mathcal{E}$ (Sec.~\ref{sec:encoder})  is an audio-visual (AV) spatial correspondence encoder that
learns an implicit representation of the spatial relationships between the visual and unmasked binaural audio tokens, while 
$\mathcal{D}$ (Sec.~\ref{sec:decoder}) is an audio-visual decoder for binaural audio inpainting that uses this implicit representation to synthesize the masked audio tokens. We also devise a simple yet novel masking protocol (Sec.~\ref{sec:masking}) 
for our inpainting task, which mixes masking random audio tokens with masking a full audio channel,
and 
helps the model 
learn stronger audio-visual spatial associations 
{that facilitate the downstream social tasks {(Sec.~\ref{sec:downstream_tasks})}.}
We train $\mathcal{F}$ 
{to} minimize the prediction error in the masked audio tokens {(Sec.~\ref{sec:training})}. Next, we 
describe our model design,
audio masking protocol, training objective and network architecture, and downstream tasks.

\subsection{Audio-visual spatial correspondence encoder}\label{sec:encoder}
The audio-visual spatial correspondence encoder $\mathcal{E}$ (Fig.~\ref{fig:model} center) extracts features from the visual and unmasked audio tokens $\big\{\hat{V}, \hat{A}^U\big\}$. It begins by embedding the visual and audio tokens using separate transformer encoders~\citep{feichtenhofer2022masked} for individually capturing the spatio-temporal features in the two modalities. Next, it uses a shared transformer encoder
to jointly encode the audio and visual features, and produces a multi-modal representation suitable for binaural audio inpainting.  
Next, we describe the individual encoders next.

\vspace*{-0.1in}
\paragraph{Video and audio encoders.} We first encode the visual tokens $\hat{V}$ using a linear layer to generate visual features $v$, such that $v = \big\{v_1, \ldots, v_P\big\}$. We encode the audio tokens $\hat{A}^U$ with another linear layer to produce audio features $a$, such that $a = \big\{a_1, \ldots, a_{Q-S}\big\}$, where $S$ is the number of masked tokens out of a total of $Q$ audio tokens (cf. Sec.~\ref{sec:task}). For each visual feature $v_j$, we add a sinusoidal positional embedding $p^V_j$~\citep{vaswani2017attention} to it, where $p^V_j$ captures cues about the 3D position of the $j^{\text{th}}$ tubelet in the visual clip $V$. For an audio feature $a_i$, we add a sinusoidal positional embedding $p^A_i$ and a learnable channel embedding $c \in \big\{c_L, c_R\big\}$ to it to convey information about the 2D location of the $i^{\text{th}}$ unmasked audio token in the spectrogram and also the audio channel to which it belongs. Next, we feed the transformed visual and audio features to separate transformer encoders, $\mathcal{E}^V$ and $\mathcal{E}^A$, respectively, and obtain visual features $e^V = \big\{e^V_1, \ldots, e^V_P \big\}$ and audio features $e^A = \big\{e^A_1, \ldots, e^A_{Q-S} \big\}$.

\vspace*{-0.05in}
\paragraph{Shared audio-visual encoder.}
Given the visual features $e^V$ and audio features $e^A$, we concatenate them into $e^{AV}$, such that $ e^{AV} = \big\{ e^V_1, \ldots, e^V_P, e^A_1, \ldots, e^A_{Q-S} \big\}$, and re-add the sinusoidal positional embeddings $p^V$ and $p^A$ to the features of the respective modalities in $e^{AV}$. Furthermore, unlike existing audio-visual masked auto-encoders~\citep{georgescu2022audiovisual, huang2022mavil, gong2023contrastive}, we add the channel embeddings $c$ to the audio features, and 
learnable modality embeddings $m \in \big\{m_A, m_V\big\}$ to all features in $e^{AV}$ to help the model distinguish between the visual and audio modalities. Next, a shared audio-visual transformer  $\mathcal{E}^{AV}$ encoder takes $e^{AV}$ as input and outputs audio-visual features $f^{AV}$, which implicitly hold spatio-temporal information required for accurate inpainting of audio.

\subsection{Audio-visual decoder for binaural audio inpainting}\label{sec:decoder}
Our  audio-visual decoder $\mathcal{D}$ (Fig.~\ref{fig:model} right) takes $f^{AV}$ as input and attempts to synthesize the masked binaural audio tokens by leveraging spatio-temporal cues 
in $f^{AV}$. 
It first projects $f^{AV}$ to a lower-dimensional feature set $g^{AV}$. It then appends a learnable embedding for the masked audio tokens to $g^{AV}$ and passes it through a shared audio-visual transformer decoder~\citep{he2022masked}. Next, it feeds the audio feature outputs of the shared decoder to another transformer decoder and uses its outputs to predict an estimate of the masked binaural audio tokens. The decoders are 
lightweight compared to the encoders, 
ensuring that the encoders are primarily
responsible for driving the inpainting task and producing good audio-visual features for strong downstream performance. We next describe each component of $\mathcal{D}$ in detail.   

\vspace*{-0.05in}
\paragraph{Shared audio-visual decoder.}
We first create a lower-dimensional projection $g^{AV}$ of the audio-visual encodings $f^{AV}$ by passing it through a linear layer, and append a learnable embedding $\phi$ corresponding to each of the $S$ masked audio tokens to $g^{AV}$. Next, we add the positional embeddings $p^V$ and $p^A$, the audio channel embeddings $c$, and the modality embeddings $m$ to $g^{AV}$, and feed it to a shallow transformer decoder $\mathcal{D}^{AV}$ that outputs an audio-visual feature set $h^{AV}$. We then take the audio features $h^A$ from $h^{AV}$ and pass them to the audio decoder for further processing. 

\vspace*{-0.05in}
\paragraph{Audio decoder.} The audio decoder $\mathcal{D}^A$ re-adds
the positional embeddings $p^A$ and channel embeddings $c$ to $g^A$, and feeds it to a
transformer decoder, which outputs audio features $d^A$.

\vspace*{-0.05in}
\paragraph{Prediction of masked audio tokens.} Finally, we take the subset 
of all audio features $d^A$
correspond{ing} to the masked audio tokens $\hat{A}^M$, upsample them by passing through a linear layer, and reshape them to obtain an estimate $\tilde{A}^M$ of the masked tokens $\hat{A}^M$, such that 
$\tilde{A}^M = \big\{\tilde{A}_1, \ldots, \tilde{A}_S \big\}$.

\subsection{Audio masking}\label{sec:masking}
Different from other masked auto-encoding counterparts~\citep{georgescu2022audiovisual,huang2022mavil,gong2023contrastive}, we design an audio masking protocol that is customized to help our model better extract spatial audio-visual cues during self-supervised pretraining.
In particular, we mix the strategy of randomly masking a full audio channel with that of randomly masking audio tokens 
{using a hyperparameter $r$} 
during training, where $r$ represents the {probability}
with which
we randomly drop a {full} audio channel {and $r$ is sampled from a uniform distribution $U(0,1)$:}
{
\begin{equation*}
\small
    \text{mask}(\hat{A}) = 
    \begin{cases}
        \hat{A}^M = A^L \, \text{or} \, \hat{A}^M = A^R & \text{if} \quad x \sim U(0,1) \leq r \\

        \hat{A}^M \subseteq \{\hat{A}_1,\dots,\hat{A}_Q\} & \text{Otherwise}
    \end{cases}
\end{equation*}
\vspace{-0.2cm}
}

On the one hand, \emph{token masking} could lead to tokens from the same location in the two audio channels being present among the unmasked tokens, 
providing additional spatial cues to the model and resulting in a simpler, {stabler} optimization objective for the inpainting task. In addition, since token masking involves masking a short span in both frequency and time domains, the model can rely more on local audio-visual cues and tolerate the global noise in both the visual and audio streams due to a camera-wearer's motion. On the other hand, \emph{channel masking} forces the model to solve a more challenging binauralization task solely on the basis of vision, which could help it learn even stronger spatial features. {This encourages the model to}
 reason about the camera motion at a more global scale (over the entire clip span).
Towards achieving high performance on the downstream tasks, we aim to strike a fine balance between these two strategies and combine the benefits of reasoning at both temporal scales.

\subsection{Training objective and network architecture}\label{sec:training}

\begin{figure}[t]
    \centering
    \begin{subfigure}[b]{0.45\textwidth}
    \centering
    \includegraphics[width=\textwidth]{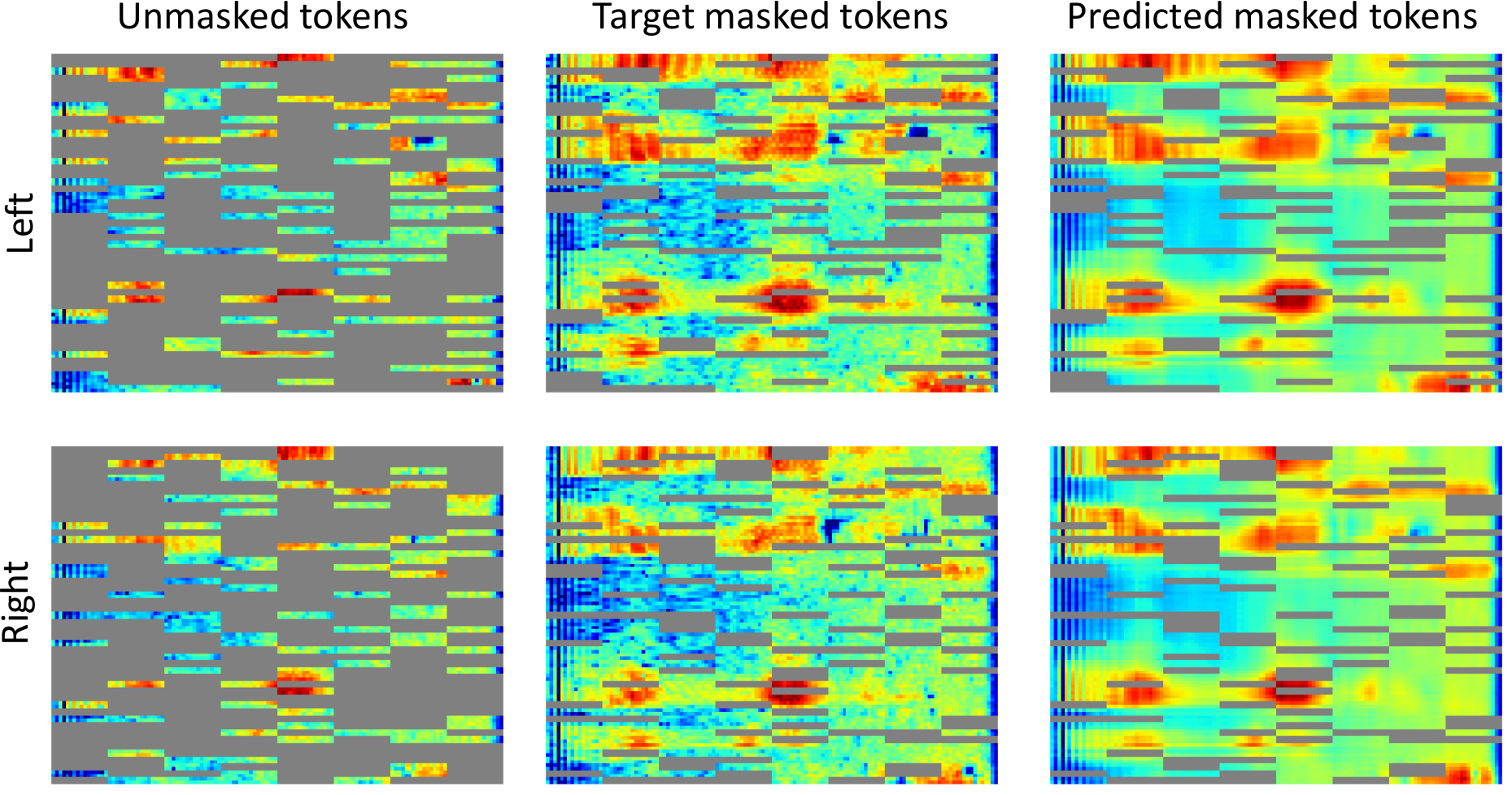}
    \caption{{Token masking}}
    \label{fig:tok_mask}
    \end{subfigure}\hfill\par\bigskip
    \vspace{-0.3cm}
    \begin{subfigure}[b]{0.45\textwidth}
    \centering
    \includegraphics[width=\textwidth]{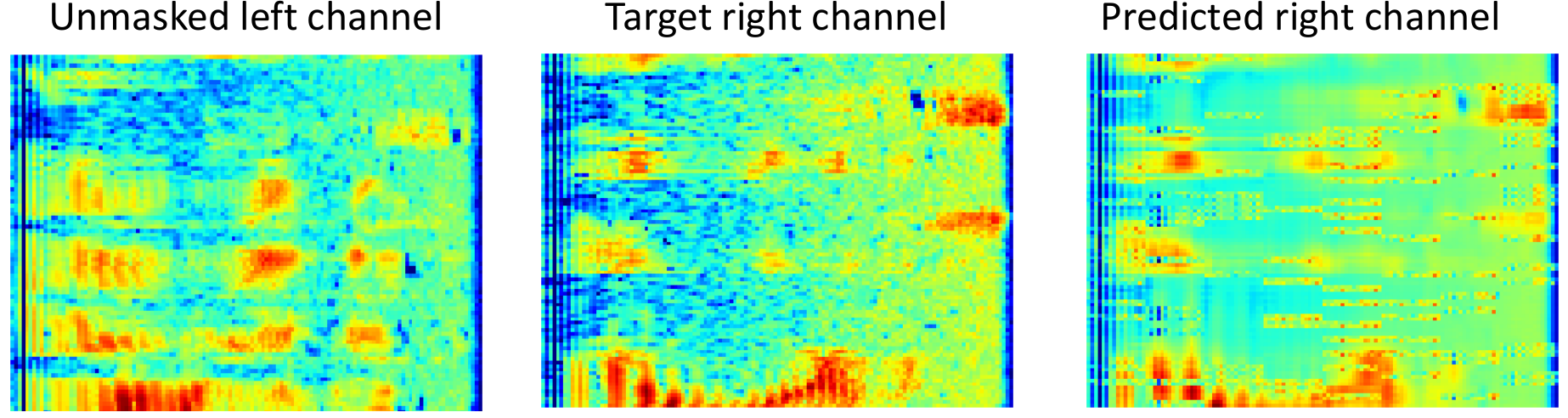}
    \caption{{Channel masking}}
    \label{fig:ch_mask}
    \end{subfigure}
    \vspace{-0.2cm}
\caption{{Masked targets and predictions shown alongside the 
    unmasked inputs for (a) token masking and (b) channel masking.
    Our predictions accurately capture the global patterns in the target spectrograms, which depend on the scene's spatial properties.}}
    \label{fig:spects_w_masks}
    \vspace{-0.1cm}
\end{figure}

We train our model to minimize the error in 
predicting the masked audio tokens. 
In particular, we compute the mean-squared error $\mathcal{L}$ averaged over all masked audio tokens:
\begin{equation}
    \mathcal{L} = \frac{1}{S}\sum_{i=1\ldots S} ||\ddot{A}_i - \tilde{a}_i ||^{2}_{2}.
\end{equation}
We visualize our predicted audio tokens in Fig.~\ref{fig:spects_w_masks} {for 
the cases of token (Fig.~\ref{fig:spects_w_masks}a) and channel (Fig~\ref{fig:spects_w_masks}b)} masking. Our model is able to accurately {predict the masked targets and} capture the global patterns in the spectrograms, which are often determined by the {spatial audio-visual cues captured} from of the scene {(visual input not shown in Fig~\ref{fig:spects_w_masks} for brevity)}, thereby further emphasizing our model’s ability to learn useful spatial features.

Our uni-modal encoders, $\mathcal{E}^A$ and $\mathcal{E}^V$,
have 8 layers, while the audio-visual encoder $\mathcal{E}^{AV}$ has 6 layers. All encoders have 12 attention heads and use 768-dimensional hidden embeddings. The audio-visual decoder $\mathcal{D}^{AV}$ and audio-only decoder $\mathcal{D}^A$ have 1 and 3 layers, respectively. Both decoders have 6 attention heads and use 384-dimensional hidden embeddings.
To pretrain our model, 
we set the relative frequency of dropping an audio channel in our masking protocol for training to $r=20 \%$ {based on disjoint validation data (see Supp.)}. 
We train our model for 200 epochs using the AdamW~\citep{loshchilov2018decoupled} optimizer with a weight decay of $10^{-5}$, and a learning rate scheduler that reaches a peak learning rate of $2 \times 10^{-4}$ over 10 warmup epochs, and then decays it through half-cycle cosine annealing~\citep{loshchilov2017sgdr}.
For data agumentation, we perform random flipping of audio channels and video clips along the frame width. 
For downstream evaluation, we fuse the features from the audio-visual encoder with the backbones for downstream tasks, and finetune them. 
See Supp. for further details on 
architecture and training.

\subsection{Downstream tasks}\label{sec:downstream_tasks}
We explore two downstream tasks with our pretrained features:  active speaker detection and spatial audio denoising.
Active speaker detection (ASD) involves matching an audio clip with an appropriate face track from the corresponding video clip, {i.e., answering ``which person is speaking now?"}. While current 
SOTA methods~\citep{tao2021someone, min2022learning} rely on semantic similarities between monaural audio and vision to solve this task, {we hypothesize that} leveraging spatial audio can additionally reveal the sound source location in the video.
In spatial audio denoising, \
the goal is to separate the target audio from distractors. 
In particular, we 
aim to remove the audio from sources extraneous to the conversation---out-of-view sounds from other parts of the scene.
{We detail the backbone models for each in the next section.}

\section{Experiments}\label{sec:experiments}
\vspace*{-0.05in}

{We validate our learned representations on two downstream tasks and  two datasets, and we compare with
prior models for spatial audio-visual feature learning~\cite{yang2020telling,gao20192,sato2022self-supervised, georgescu2022audiovisual}, as well as various baselines and ablations.}

\vspace{-0.2cm}
\paragraph{Datasets.} We train and evaluate our model on two challenging egocentric datasets 
{containing}
video and binaural audio {of people having conversations}: \textbf{1)} EgoCom~\citep{9200754}, and \textbf{2)} EasyCom~\citep{donley2021easycom}, detailed in Supp.  
{To our knowledge, these are the only two publicly available datasets offering binaural audio with conversations in video, whether exocentric or egocentric.  In particular, Ego4D~\cite{grauman2022ego4d} and EPIC~\cite{Damen2022RESCALING} 
do not comprise social scenarios and
are not applicable.}
Whereas EasyCom primarily 
{has}
participants sitting around a table and talking, EgoCom has videos of participants moving around a room, turning their face and body, standing up, etc. 
{They}
test 
our method{'s} robustness in diverse scenarios of varying difficulty. {See Supp. for more details.}

\subsection{Active speaker detection}\label{sec:main_asd}
We first evaluate 
on active speaker detection (ASD).
\vspace*{-0.05in}

\vspace{-0.1cm}
\paragraph{Backbone models.} 
We consider two 
SOTA ASD models as the backbones for leveraging 
our pretrained representations: \textbf{1)} TalkNet~\citep{tao2021someone}, and \textbf{2)} SPELL~\citep{min2022learning}.  
TalkNet 
encodes
a face track and an audio clip using attention for learning intra- and inter-modal semantic and temporal features. Next, it fuses these features and performs binary classification to predict if the face in the track is active.
SPELL 
extracts audio-visual features for each face in a clip using 
ResNets~\citep{Alcazar2020ActiveSI},
and learns long and short-term semantic relations among them using a graph neural network. Finally, it performs binary classification of these features for predicting active speakers.

\vspace{-0.1in}
\paragraph{Pretrained feature fusion.} 
To fuse our pretrained features with the ASD backbones, we  use a transformer decoder that cross-attends to the feature outputs of the shared encoder $\mathcal{E}^{AV}$ using sinsuoidal embedding as queries, with each embedding representing a clip frame index.
Next, we append
the decoder outputs to the cross-attention outputs in TalkNet, or the 
audio-visual encoder outputs in SPELL, 
frame by frame.
In essence, while the original audio-visual encoders leverage \emph{semantic} correlations between vision and audio, our features provide strong complementary \emph{spatial} cues.

\vspace*{-0.05in}
\paragraph{Baselines.} For both TalkNet and SPELL, we compare against multiple baselines comprising both the unmodified backbone and improved versions of it, in addition to some naive methods: 
\begin{itemize}[leftmargin=*,topsep=0pt,partopsep=0pt,itemsep=0pt,parsep=0pt]
\item \textbf{All-active:} a naive model that predicts that all visible
faces are always active 
\item \textbf{All-inactive:} a naive model that predicts that all visible faces are always inactive
\item \textbf{Random:} a naive model that emits a random ASD confidence score for every visible speaker 
\item \textbf{Backbone w/o audio:} a vision-only version of the backbone with no audio input
\item \textbf{Backbone:} the originally-proposed backbone that processes only faces and monaural audio
\item \textbf{Backbone-binaural:} an improved backbone
using binaural audio instead of monaural, alongside positional encodings for the faces, indicative of their relative position and depth, for better matching the face to the audio
\item \textbf{Backbone-binaural w/ scene:}
a further improvement over the backbone, where we 
also provide the scene images (uncropped video frames) to backbone-binaural
\item \textbf{Backbone w/ TLR~\citep{yang2020telling}:} 
fuses features from the SOTA Telling Left from Right (TLR)~\citep{yang2020telling}, which learns audio-visual spatial correspondences by predicting the spatial alignment between vision and binaural audio
\item \textbf{Backbone w/ 2.5D-VS~\citep{gao20192}:}
fuses features from the SOTA audio-visual binauralization model, 2.5D Visual Sounds (2.5D-VS)~\citep{gao20192}
\item \textbf{Backbone w/ 2.5D-VS~\citep{gao20192}++:} fuses features from 2.5D-VS with a transformer architecture
\item \textbf{Backbone w/ SSR~\citep{sato2022self-supervised}{++}:} 
fuses features from the SOTA self-supervised audio-visual stereo sound ranking (SSR)~\citep{sato2022self-supervised} model {with a transformer architecture}
\item \textbf{Backbone w/ AV-MAE~\citep{georgescu2022audiovisual}:}
fuses features from the SOTA modality-inpainting AV-MAE~\citep{georgescu2022audiovisual} model
\end{itemize}
For all alternate feature-learning methods~\cite{georgescu2022audiovisual,sato2022self-supervised,gao20192,yang2020telling}, we pretrain them on our datasets and use 
{our} feature fusion method.
{Thus, any advantages in performance of our approach over these SOTA representations will be attributable to our modeling ideas.} 
{Importantly}, the {2.5D-VS~\citep{gao20192}++, SSR~\cite{sato2022self-supervised}{++}, and AV-MAE~\cite{georgescu2022audiovisual} features 
{all rely on transformers and} have similar model capacity as ours (see Supp.~for a detailed analysis on model capacity).
We use the standard \textbf{mean average precision} (mAP) metric.

\begin{table}[!t]
\small
  \centering
  \setlength{\tabcolsep}{2pt}
  \resizebox{\linewidth}{!}{
    \begin{tabular}{l c c c c}
    \toprule
     &  \multicolumn{2}{ c| }{TalkNet~\citep{tao2021someone}} & \multicolumn{2}{ c }{SPELL~\citep{min2022learning}} \\
    Model &  \multicolumn{1}{ c }{\textit{EgoCom}} & \multicolumn{1}{ c| }{\textit{EasyCom}} & \multicolumn{1}{ c }{\textit{EgoCom}} & \multicolumn{1}{ c}{\textit{EasyCom}}\\
    \midrule
    \multicolumn{1}{ l }{\textit{No pretraining}}\\
     \quad All-active  & 32.9 & 30.1 & 32.9 & 30.1 \\
     \quad All-inactive  & 32.9 & 30.1 & 32.9 & 30.1 \\
     \quad Random  & 30.8 & 28.0 & 30.8 & 28.0  \\
     \quad B w/o audio  & 41.5 & 50.1 & 60.4 & 63.2 \\
     \quad B & 52.8 & 45.7 & 60.9 & 59.0  \\
     \quad B-binaural & 60.0 & 59.6 & 63.1 & 60.3 \\
     \quad B-binaural w/ scene   & 60.8 & 66.9 & 61.2 & 61.4 \\
     \multicolumn{1}{ l }{\textit{Alternate pretraining methods}}\\
     \quad B w/ TLR~\citep{yang2020telling}  & 47.9 & 59.3 & 61.3 & 61.7 \\
     \quad B w/ 2.5D-VS~\citep{gao20192}  & 57.7 & 63.7 & 61.2 & 59.7 \\
     \quad B w/ 2.5D-VS~\citep{gao20192}++  & 63.4 & 68.3 & 65.1 & 64.5 \\
     \quad B w/ SSR
     ~\citep{sato2022self-supervised}{++}  & 61.2& 70.6 & 61.2 & 67.4\\
     \quad B w/
     AV-MAE~\citep{georgescu2022audiovisual}  & 61.1 &  61.3 &  64.4 & 65.2\\
    \textbf{Ours} & \textbf{63.9} & \textbf{71.8} & \textbf{65.6} & \textbf{70.2}\\
    \midrule
    Ours w/o pretrain & 62.7 & 62.9 & \_\_  & \_\_ \\
    Ours w/ pretrain monaural & 61.0 & 69.4 & 63.9 & 69.0\\
    \bottomrule
  \end{tabular}
  }
  \caption{Mean average precision ($\%$) for active speaker detection with TalkNet~\citep{tao2021someone} and SPELL~\citep{min2022learning} backbones {on both datasets}. Higher is better. {`B' refers to backbone.}  Note that SPELL requires storing pretrained features in the graph nodes; therefore it does not allow training from scratch.
  }
  \label{tab:asd_main}
  \vspace*{-0.1in}
\end{table}

 \vspace{-0.1in}
\paragraph{Results.}
Table~\ref{tab:asd_main} (top) reports our ASD results on both val and test splits. The three naive baselines {perform poorly} on both EgoCom~\citep{9200754} and EasyCom~\citep{donley2021easycom}, emphasizing the difficulty of the task.
For both TalkNet~\citep{tao2021someone} and SPELL~\citep{min2022learning}, the unchanged backbone model generally performs better than the model without audio, showing that both vision and audio are required.
Upgrading from monaural to binaural audio further boosts performance, as the model can now leverage both spatial and semantic information. Additionally using scene features lets the backbone explicitly match the scene area around the inferred source location with the face, and further improves ASD, especially for EgoCom, where the background scene changes more often. 

Among alternate {feature learning}
methods, 2.5D-VS~\citep{gao20192}++, SSR~\citep{sato2022self-supervised}{++} and AV-MAE~\citep{georgescu2022audiovisual} consistently outperform TLR~\citep{yang2020telling} and 2.5D-VS~\citep{gan2019self}, and also the basic and enhanced backbones, showing that self-attention and higher model capacity 
{enhance feature quality.}
{Besides,}
2.5D-VS outperforms TLR, and 2.5D-VS++ and AV-MAE 
generally outperform
SSR{++}, {indicating that objectives that {promote reasoning} directly at the spectrogram level 
{improve}
results.} 

Our model substantially outperforms all baselines---{including the SOTA AV representation learning methods---}for both backbones (TalkNet and SPELL) on both datasets. This shows that our method 
 learns stronger spatial features
 that are both backbone- and dataset-agnostic. 
In contrast, methods developed for exocentric settings  {with more stationary cameras} (such as TLR and 2.5D-VS) rely more on the global visual context and seem to struggle in our setting, {where the camera moves frequently and the sound source leaves the field of view.}
Finally, our improvement over {2.5D-VS++,} SSR{++} and AV-MAE, which use similar encoders as ours, disentangles the 
benefits of our masking strategy and model design from those of 
the model capacity.

\vspace{-0.1in}
\paragraph{Model analysis.}
Table~\ref{tab:asd_main} (bottom) shows 
ablations
of our 
method. Upon training for ASD from scratch, we see sharp drop in performance,
showing that our advantage is not solely due to our model design, but also our self-supervised pretraining stage. Performance also declines upon pretraining with monaural audio instead of binaural, showing that our model goes beyond learning semantic features and successfully captures spatial features useful for ASD.

\subsection{Spatial audio denoising}\label{sec:main_sad}

\begin{table}[!t]
\small
  \centering
  \setlength{\tabcolsep}{2pt}
    \begin{tabular}{lrr}
    \toprule
    Model & SI-SDRi $\uparrow$ & STFT $\downarrow$\\
    \midrule
    \multicolumn{1}{ l }{\textit{No pretraining}}\\
    \quad B w/o vision & 1.61 & 7.36 \\
    \quad B & 1.46 & 7.27 \\
    \quad B w/ ImageNet features & 1.48 & 6.95 \\
     \multicolumn{1}{ l }{\textit{Alternate pretraining methods}}\\
     \quad B w/ TLR~\citep{yang2020telling}   & 1.41 & 7.79 \\
     \quad B w/ 2.5D-VS~\citep{gao20192}   & 1.67 & 7.34 \\
     \quad B w/ 2.5D-VS~\citep{gao20192}++  & 2.11 & 6.60 \\
     \quad B w/ SSR~\citep{sato2022self-supervised}
     {++} 
     & 2.04 & 6.70\\
     \quad B w/ 
     AV-MAE~\citep{georgescu2022audiovisual} & 2.07 & 6.62\\
    \textbf{Ours} & \textbf{2.20}& \textbf{6.51}\\
    \midrule
    B w/o pretrain & 1.90 & 7.25\\
    B w/ pretrain monaural audio & 2.00 & 6.75\\
    \bottomrule
  \end{tabular}
  \caption{Audio denoising with U-Net~\citep{yang2020telling} backbone (referred to as `B' in table) for 
  0 dB noise (maximum). {See Supp.~for varying noise levels.}
  All STFT distance measures use base $10^{-3}$.
  }
  \label{tab:sep_main}
  \vspace*{-0.1in}
\end{table}

Next we evaluate spatial audio denoising {on EgoCom}.\footnote{
For EasyCom, the task setup is
 {ill-posed} 
for all {models}
because mixing audio from a different EasyCom clip as noise
leads to spatially overlapping sound sources, since all clips in the dataset 
are recorded 
at the same physical location
(people sitting around 
the same table in the same room).  
}
To instantiate this task, we add the binaural audio of a target clip with the downscaled binaural audio from another randomly chosen clip, where the downscaling factor depends on the desired noise level.  {The goal is to} 
extract the target {sound} from the mixture. We evaluate three noise levels, expressed using the signal-to-noise (SNR) ratio: \textbf{1)} 0 dB, \textbf{2)} 2.5 dB, and \textbf{3)} 5 dB. 
The different noise levels test our model's robustness to varying levels of task difficulty---the lower the SNR value, the higher the noise and difficulty.

\vspace*{-0.05in}
\paragraph{Backbone model.} We adopt the commonly used U-Net~\citep{ronneberger2015u} for audio-visual source separation~\citep{yang2020telling, gao20192} as the backbone, which produces a binaural ratio mask for the target audio (see Supp.~for details). 
We multiply the predicted ratio mask with the mixed magnitude spectrogram to get the predicted magnitude spectrogram, then convert it to a waveform using 
inverse short-time Fourier transform with the mixed audio phase. 

\vspace*{-0.05in}

\paragraph{Pretrained feature fusion.}
To use our features for denoising, we reshape the visual features 
and unmasked audio features
produced by our audio-visual encoder $\mathcal{E}^{AV}$ to form multi-channel 2D maps, where the features align with their corresponding tokens vis-a-vis the raster order. Next, we pass the feature maps to separate convolutional layers, concatenate the outputs channel-wise, and use them to replace the visual features at the U-Net~\citep{yang2020telling} bottleneck.

\vspace*{-0.05in}

\paragraph{Baselines.}
We compare against the following baselines and existing methods:
\begin{itemize}[leftmargin=*,topsep=0pt,partopsep=0pt,itemsep=0pt,parsep=0pt]
    \item \textbf{U-Net w/o vision:} an audio-only blind denoising model
    \item \textbf{U-Net:} the original backbone without any alterations
    \item \textbf{U-Net w/ ImageNet:} 
    pretrains the visual encoder on ImageNet~\citep{5206848}
    \item \textbf{U-Net w/ TLR~\citep{yang2020telling}:} 
    fuses 
    features from TLR~\citep{yang2020telling} 
    \item \textbf{U-Net w/ 2.5D-VS~\citep{gao20192}:} 
    fuses 
    pretrained features from 2.5D-VS~\citep{gao20192} 
    \item \textbf{U-Net w/ 2.5D-VS~\citep{gao20192}++:} 
    fuses features from the transformer-based version of 2.5D-VS
    \item \textbf{U-Net w/ SSR~\citep{sato2022self-supervised}{++}:} 
    fuses features from {the transformer-based version of} SSR~\citep{sato2022self-supervised} 
    \item \textbf{U-Net w/ AV-MAE~\citep{sato2022self-supervised}:} 
    fuses features from the modality inpainting AV-MAE~\citep{georgescu2022audiovisual} model
\end{itemize}

\paragraph{Evaluation metric.} For evaluating our denoising quality, we use standard metrics: \textbf{1) STFT} distance
(the L2 error between the predicted and ground-truth spectrograms)
expressed using base $10^{-3}$ and \textbf{2) SI-SDRi}: the improvement in SI-SDR~\citep{LeRoux2018SDRH}, a scale-invariant estimate of the level distortion in the audio, over using the mixed audio as the prediction.

\begin{figure}[!t] 
    \centering
    \includegraphics[width=1.\linewidth]{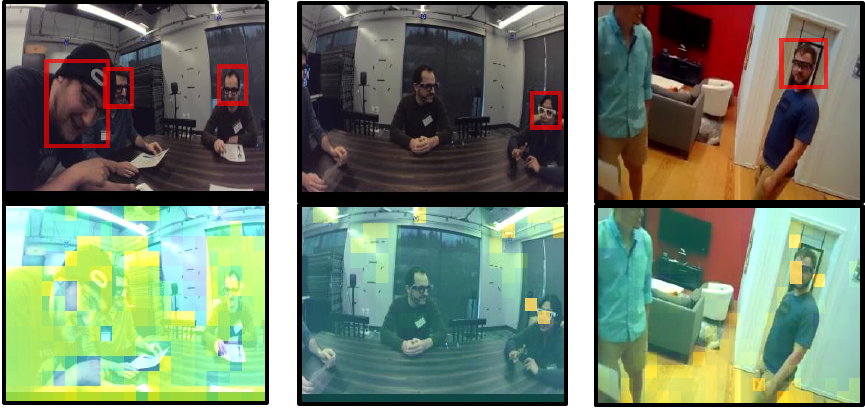}
    \caption{Heat maps showing the image areas our 
    model's AV encoder attends to, placed alongside the images. 
    {Brighter yellow means higher attention score.}
    Our model attends to image regions (\eg faces of speakers, 
    {{sound-reflecting} flat regions like }{floor and table,} 
    etc.) 
    that strongly determine the spatial properties of the audio, including direct sources of sound (marked in red). 
    }
\label{fig:qual}
\end{figure}

\paragraph{Results.}
Table~\ref{tab:sep_main} (top) shows spatial audio denoising results on the challenging EgoCom dataset with 0 dB, the most difficult noise level.  {See Supp.~for similar results with the other noise levels.} 
The 
unchanged U-Net backbone
lowers the STFT distance compared to
the version that lacks vision,
showing
that 
like ASD, vision is crucial for better denoising. 
Using pretrained features of 
2.5D-VS~\cite{gao20192} (++),
SSR~\citep{sato2022self-supervised}{++} or AV-MAE~\citep{georgescu2022audiovisual} further improves the performance, showing that learning spatial audio-visual features aids denoising.

Our method outperforms all baselines ($p \leq 0.05$) across both metrics. 
While the improvement over the baselines that do not use self-supervised pretraining emphasizes the utility of learning spatial audio-visual relationships through self-supervision, the 
{gain}
over 
other pretraining methods
underlines the strengths of our self-supervised method design---which are consistently realized for both ASD and denoising. 
Further, our 
{performance margins} 
are larger for higher noise levels (0 and 2.5 dB),
indicating that our features play a bigger role in the more difficult denoising settings. 

\vspace{-0.25cm}
\paragraph{Model analysis.}
In Table~\ref{tab:sep_main} (bottom), we ablate our pretraining method.
Similar to ASD, training 
from scratch on the denoising task 
hurts performance. This disentangles the impact of our pretext task design from the model architecture and shows that our pretraining stage helps the backbone with learning better audio-visual features, leading to superior denoising quality.  Furthermore, pretraining with monaural audio also degrades performance, re-emphasizing that our method is not  restricted to learning semantic features{---in contrast to prior work~\cite{georgescu2022audiovisual,huang2022mavil,gong2023contrastive}}.

See Supp. for additional analysis of the effect of alternate masking choices, multi-level positional embeddings, tasks-specific backbones, and our finetuning strategy on
performance.

\begin{figure}[!t]
    \centering
    \begin{subfigure}[b]{0.5\textwidth}
    \centering
    \includegraphics[width=\textwidth]{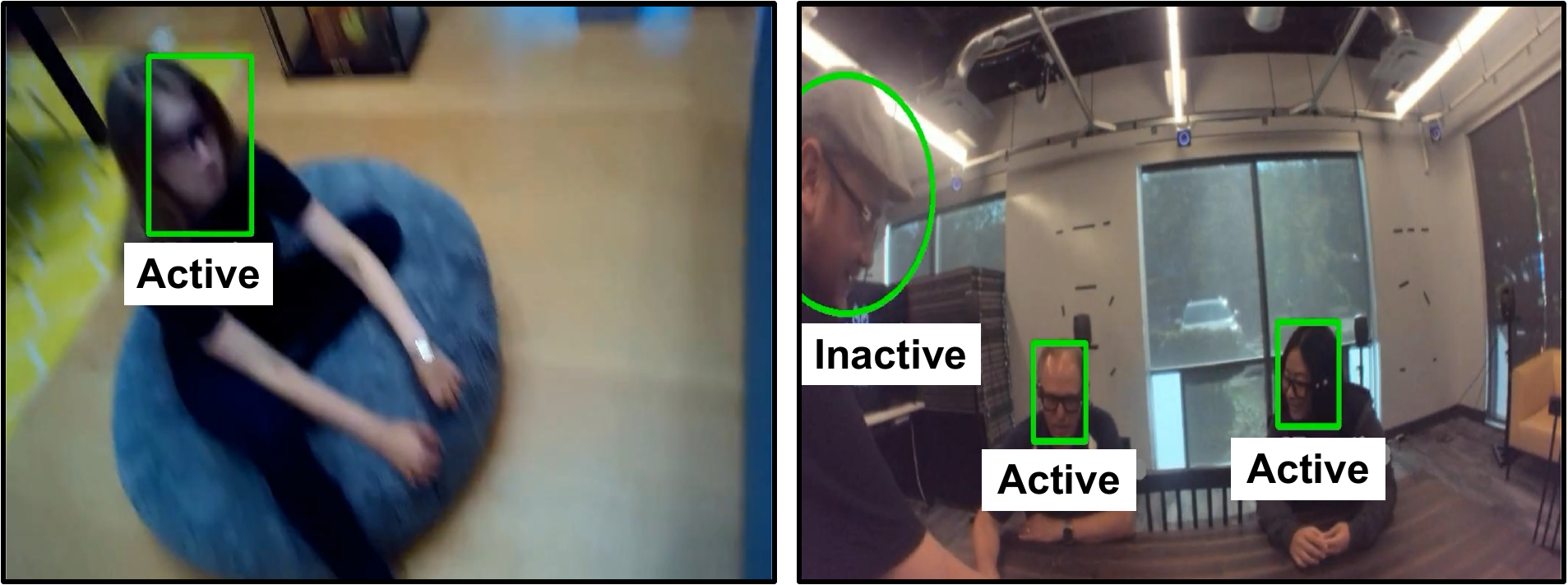}
    \caption{{Our model succeeds on ASD even with fast camera motion (left; note image blur), multiple active speakers (right) and partially visible faces (left and right)}}
    \label{fig:success_ASD}
    \end{subfigure}\hfill\par\bigskip
    \vspace{-0.3cm}
    \begin{subfigure}[b]{0.5\textwidth}
    \centering
    \includegraphics[width=\textwidth]{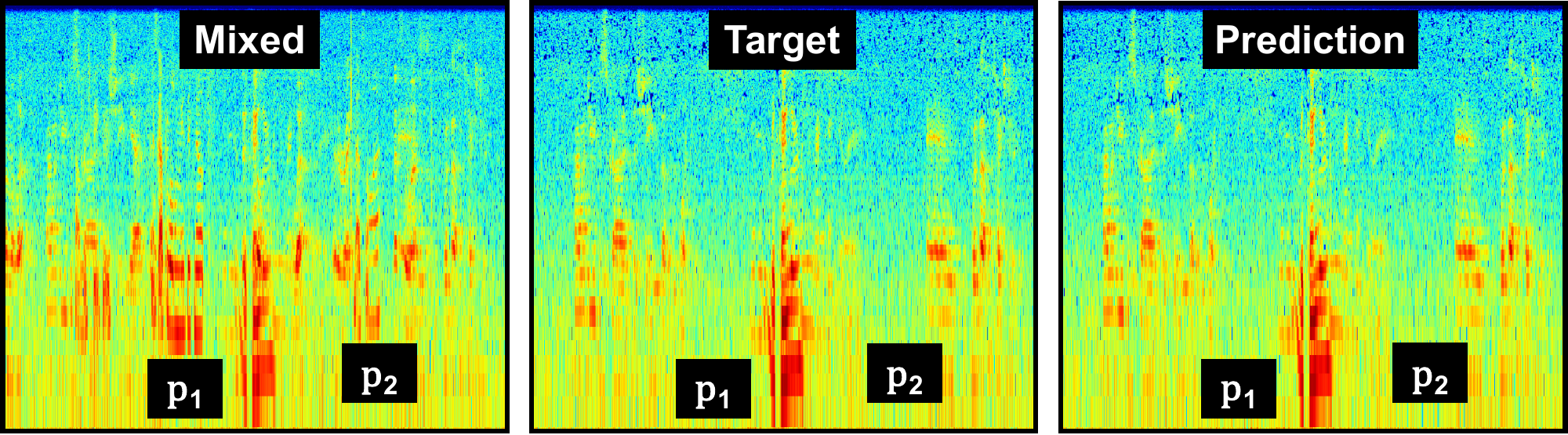}
    \caption{{Our model denoises accurately---note the noise patches in mixed audio above points $p_1$ and $p_2$, which are successfully removed in our prediction.}}
    \label{fig:success_SAD}
    \end{subfigure}
    \vspace{-0.2in}
\caption{{Success cases for ASD (a) and denoising (b)}}
    \label{fig:success_down}
    \vspace{-0.4cm}
\end{figure}

\subsection{Qualitative analysis}\label{sec:qual_analysis}
In Fig.~\ref{fig:qual}, we analyze the visual attention maps of our shared 
encoder $\mathcal{E}^{AV}$. 
{Note} that the regions of high attention are not only limited to the 
{direct sound sources} (\eg, 
{regions in and around}
faces of active speakers across examples),
but also include large 
{sound-reflecting}
objects (\eg, 
{the flat surface of the }table on the left; the walls on the left and in the middle;
the floor
on the right, etc.) that determine how sound spatializes through early reflections, late reverberations, etc.
Interestingly, our model also attends to multiple people if they are speaking at the same time (see left), thereby facilitating the detection of multiple active speakers. See Supp.~for additional visualizations showing 
how, depending on the scene's spatial layout, our model uses one audio channel more than the other to attend to important image locations.

{In Fig.~\ref{fig:success_down}, we qualitatively show our model's success cases. On ASD (Fig.~\ref{fig:success_ASD}), our model can tackle drastic camera movements, multiple active speakers, and partially visible faces. On denoising (Fig.~\ref{fig:success_SAD}), our model is able to remove interferences from distractor sources, and make predictions that closely match the ground truth in spectrogram structure.}

{We also observe some limitations.}  Our model's performance on ASD declines when there are drastic movements of the camera wearer, or there is a high overlap in speech from different conversation participants. On denoising, our model struggles when the noisy audio is semantically and acoustically similar to the target, or when it cannot extract spatial cues due to occlusions or out-of-view speakers. Refer to our Supp.~video for both success and failure cases.

\section{Conclusion}\label{sec:conclusion}

We {introduce}
a novel self-supervised approach for 
learning audio-visual representations in social egocentric videos via spatial correspondence between the video and its binaural audio. 
Through extensive evaluation, we show that our learned features are strong and generic enough to improve over multiple backbone methods 
{on}
multiple downstream tasks. 
{In future work, we 
will explore how the learned spatial audio-visual cues may reveal the social attention between speakers.}

\small{\noindent \textbf{Acknowledgements:} UT Austin is supported in part by NSF CCRI and the IFML NSF AI Institute. KG is paid as a research scientist by Meta, and SM is a visiting researcher at Meta.}

{
    \small
    \bibliographystyle{ieeenat_fullname}
    \bibliography{mybib}
}

\clearpage
\section{Supplementary Material}
In this supplementary material, we provide additional details about:
\begin{itemize}
    \item Video (with audio) for qualitative illustration of our pretext task and qualitative evaluation of our model 
    on the downstream tasks (Sec.~\ref{sec:supp_vid}){, as noted in Sec.~\ref{sec:qual_analysis} in main}.
    \item {Spatial audio denoising results with varying noise levels (Sec.~\ref{sec:supp_sad}), as mentioned in Sec.~\ref{{sec:main_sad}} in main}
    \item Evaluation of the impact of the channel masking probability $r$ (from  
    {Sec.~\ref{sec:masking} and~\ref{sec:training}}
    in main) in our audio masking protocol 
    (Sec.~\ref{sec:r_val})
    \item {Analysis of the effect of alternate audio masking choices
    (Sec.~\ref{sec:alt_aud_masking}), as referenced  
    {in Sec.~\ref{sec:main_sad}} in main}
    \item Study of the effect of multi-level positional embeddings (Sec.~\ref{sec:multiLevel_PEs}), as noted in Sec.~\ref{sec:main_sad} in main
    \item Analysis of the effect of task-specific backbones (Sec.~\ref{sec:taskSpecific_backbones}),
as mentioned in Sec.~\ref{sec:main_sad} in main
    \item {Comparison of model capacity and computational cost among different pretraining methods (Sec.~\ref{sec:model_capacity}), as noted in Sec.~\ref{sec:main_sad} in main}
    \item {Qualitative analysis of the visual attention maps for left and right audio channels separately (Sec.~\ref{sec:attn_map_per_ch}), as referenced in Sec.~\ref{sec:qual_analysis} in main}
    \item {Analysis of the impact of our finetuning strategy (Sec.~\ref{sec:finetune_strategy}), as noted after Sec.~\ref{sec:main_sad} in main}
    \item Evaluation of the impact of our model parameter initialization on the downstream performance (Sec.~\ref{sec:param_init})
    \item Additional dataset details (Sec.~\ref{sec:supp_data}), as mentioned in Sec.~\ref{sec:experiments} 
    in main
    \item Additional model architecture and hyperparameter details for both self-supervised pretraining and downstream training (Sec.~\ref{sec:modelArc_n_training}), as referenced in Sec.~\ref{sec:training} 
    in main
\end{itemize}

\subsection{Supplementary video}\label{sec:supp_vid}
The supplementary video provides a qualitative illustration of our pretraining task for learning spatial features from audio-visual correspondence in egocentric videos{, and our proposed approach}. Moreover, we provide video samples from the both EgoCom~\citep{9200754} and EasyCom~\citep{donley2021easycom} datasets to illustrate the unique challenges posed by the egocentric videos. Additionally, we demonstrate our model's prediction quality for both active speaker detection and spatial audio denoising, and analyze common failure models for our model on both tasks. 
The video is available on \url{http://vision.cs.utexas.edu/projects/ego_av_corr}.

\subsection{Denoising with varying noise levels}\label{sec:supp_sad}
\begin{table*}[!t]
\small
  \centering
  \setlength{\tabcolsep}{2pt}
    \begin{tabular}{lrr|rr}
    \toprule
      & \multicolumn{2}{ c| }{2.5 dB} & \multicolumn{2}{ c}{5 dB}\\
    Model & SI-SDRi $\uparrow$ & STFT $\downarrow$ & SI-SDRi $\uparrow$ & STFT $\downarrow$\\
    \midrule
    \multicolumn{1}{ l }{\textit{No pretraining}}\\
    \quad U-Net w/o vision & 1.91 & 5.32 & 2.02 & 3.04\\
    \quad U-Net & 2.04 & 4.72 & 2.05 & 2.85\\
    \quad U-Net w/ ImageNet features & 2.04 & 4.66 & 2.24 & 2.74\\
     \multicolumn{1}{ l }{\textit{Alternate pretraining methods}}\\
     \quad U-Net w/ TLR~\citep{yang2020telling} features & 1.70 & 5.40 & 2.00 & 2.77\\
     \quad U-Net w/ 2.5D-VS~\citep{gao20192} features & 1.81 & 4.81 & 2.22 & 2.62\\ 
     \quad U-Net w/ 2.5D-VS~\citep{gao20192}++ features & 2.65 & 4.31 & \textbf{2.79} & \textbf{2.48}\\
     \quad U-Net w/ SSR~\citep{sato2022self-supervised}++ 
     features & 2.25 & 4.63 & 2.21 & 2.80 \\
     \quad U-Net w/ 
     AV-MAE~\citep{georgescu2022audiovisual} features & 2.46 & 4.60 & 2.14 & 2.93\\
    \textbf{Ours} & \textbf{2.72} &\textbf{4.22} & 2.46 & 2.70\\
    \midrule
    Ours w/o pretraining & 2.30 & 4.54 & 2.15 & 2.83\\
    Ours w/ pretraining using monaural audio & 2.58 & 4.38 & 2.31 & 2.81\\
    \bottomrule
  \end{tabular}
  \caption{Audio denoising with U-Net~\citep{yang2020telling} backbone for 
  {varying}
  noise levels. All STFT distance measures use base $10^{-3}$.
  }
  \label{tab:sep_supp}
  \vspace*{-0.1in}
\end{table*}

{In table~\ref{tab:sep_main} in main, we evaluated denoising with 0 dB noise. Here, we analyze the effect of varying the noise level. Table~\ref{tab:sep_supp} reports the results with 2.5 dB and 5 dB noise. We observe general similarity in performance trends across all noise levels. Whereas our model outperforms the baselines in the high-noise settings (0 and 2.5 dB), using 2.5D-VS~\citep{gao20192}++ improves the separation quality for 5 dB, underlining that our features are especially important for tackling the more challenging high-noise settings.}

\subsection{Channel masking {probability} $r$}\label{sec:r_val}

\begin{table}[h]
  \centering
  \setlength{\tabcolsep}{4pt}
    \begin{tabular}{lrr|rr}
    \toprule
     &  \multicolumn{2}{ c| }{TalkNet~\citep{tao2021someone}} & \multicolumn{2}{ c }{SPELL~\citep{min2022learning}} \\
    $r (\%)$ & Val & Test & Val & Test\\
    \midrule
    0 & 67.9 & 62.9 & 67.6 & 65.3\\
    \textbf{20 (Ours)} & \textbf{68.7} & \textbf{63.9}& \textbf{68.4} & \textbf{65.6} \\
    50 & 64.5 & 63.1 & 67.5 & 65.2 \\
    80 & 64.4 & 61.8 & 64.7 & 60.1\\
    100 & 67.9 & 63.4 & 66.1 & 65.1\\
    \bottomrule
  \end{tabular}
  \caption{Effect of $r$ on the mean average precision ($\%$) of our model for active speaker detection with two different backbones (TalkNet and SPELL).}
  \label{tab:asd_r_val}
\end{table}

\begin{table*}[!thb]
\small
  \centering
  \setlength{\tabcolsep}{2pt}
    \begin{tabular}{lrr|rr|rr}
    \toprule
     &  \multicolumn{2}{ c| }{0 dB} & \multicolumn{2}{ c| }{2.5 dB} & \multicolumn{2}{ c}{5 dB}\\
    $r(\%)$ & SI-SDRi $\uparrow$ & STFT $\downarrow$ & SI-SDRi $\uparrow$ & STFT $\downarrow$ & SI-SDRi $\uparrow$ & STFT $\downarrow$\\
    \midrule
    0 & 2.17 & 6.60 & 2.57 & 4.38 & \textbf{2.85} & \textbf{2.35} \\
    \textbf{20 (Ours)} & \textbf{2.20}& \textbf{6.51} & \textbf{2.72} &\textbf{4.22} & 2.46 & 2.70\\
    50 & 1.92 & 7.19 & 2.30 & 4.60 & 2.09 & 2.80\\
    80 & 1.82 & 7.55 & 1.98 & 5.05 & 1.68 & 3.30\\
    100 & 2.11 & 6.60 & 2.65 & 4.31 & 2.79 & 2.48 \\
    \bottomrule
  \end{tabular}
  \caption{Effect of $r$ on our model performance for audio denoising.}
  \label{tab:sep_r_val}
\end{table*}

Here, we analyze the effect of the channel masking 
{probability} $r$ in our audio masking protocol 
(Sec.~\ref{sec:masking} in main) 
on the downstream task performance. Table~\ref{tab:asd_r_val} reports the active speaker detection (ASD) results on the more challenging EgoCom~\citep{9200754} dataset, and table~\ref{tab:sep_r_val} reports the denoising results for different noise levels. We notice that the performance on both ASD and denoising, especially at the higher noise levels, declines upon increasing or decreasing the value of $r$ from our choice of 20 $\%$ based on the downstream validation performance
(Sec.~\ref{sec:training} in main), which helps our model achieve a fine balance between the two complementary strategies of masking a complete channel and randomly masking audio tokens. Whereas randomly masking a channel of the binaural audio entails solving the more under-constrained and consequently complex binauralization task, thereby helping our model learn stronger spatial associations between vision and audio, randomly masking audio tokens helps with improving training stability.

\subsection{Alternate audio masking choices}\label{sec:alt_aud_masking}
\begin{table*}[!t]
\small
  \centering
  \setlength{\tabcolsep}{2pt}
    \begin{tabular}{lrr|rr|rr|rr}
    \toprule
     &  \multicolumn{4}{ c| }{TalkNet~\citep{tao2021someone}} & \multicolumn{4}{ c }{SPELL~\citep{min2022learning}} \\
     &  \multicolumn{2}{ c| }{\textit{EgoCom}} & \multicolumn{2}{ c| }{\textit{EasyCom}} & \multicolumn{2}{ c| }{\textit{EgoCom}} & \multicolumn{2}{ c}{\textit{EasyCom}}\\
    Model & Val & Test & Val & Test & Val & Test & Val & Test\\
    \midrule
    Ours w/ time masking & 63.3 & 59.3 & \textbf{60.9} & 66.4 & 64.1 & 60.7 & 68.5 & 43.5\\
    Ours w/ frequency masking & 64.1 & 62.1 & 59.2 & 70.9 & 67.6 & 63.2 & 68.7 & 69.4\\
    Ours w/ time-frequency masking & 65.4 & 63.1 & 56.3 & 63.3 & 67.5 & 65.1 & 68.6 & 69.1\\
    \textbf{Ours} & \textbf{68.7} & \textbf{63.9} & 60.5 & \textbf{71.8} & \textbf{68.4} & \textbf{65.6} & \textbf{68.9} & \textbf{70.2}\\
    \bottomrule
  \end{tabular}
  \caption{{ASD with our model when pretrained with other audio masking choices~\citep{huang2022masked}.}}
  \label{tab:asd_altAudMask}
\end{table*}

\begin{table*}[!t]
\small
  \centering
  \setlength{\tabcolsep}{2pt}
    \begin{tabular}{lrr|rr|rr}
    \toprule
     &  \multicolumn{2}{ c| }{0 dB} & \multicolumn{2}{ c| }{2.5 dB} & \multicolumn{2}{ c}{5 dB}\\
    Model & SI-SDRi $\uparrow$ & STFT $\downarrow$ & SI-SDRi $\uparrow$ & STFT $\downarrow$ & SI-SDRi $\uparrow$ & STFT $\downarrow$\\
    \midrule
    Ours w/ time masking & 1.82 & 7.41 & 1.98 & 4.88 & 2.07 & 2.80\\
    Ours w/ frequency masking & 2.05 & 7.04 & 2.33 & 4.85 & 2.25 & 2.79\\
    Ours w/ time-frequency masking & 1.91 & 7.12 & 2.14 & 5.15 & 1.81 & 3.13\\
    \textbf{Ours} & \textbf{2.20}& \textbf{6.51} & \textbf{2.72} &\textbf{4.22} & \textbf{2.46} & \textbf{2.70}\\ 
    \bottomrule
  \end{tabular}
  \caption{{Denoising with our model when pretrained with other audio masking choices~\citep{huang2022masked}. All STFT distance measures use base $10^{-3}$.}}
  \label{tab:sep_altAudMask}
\end{table*}

{Here, we evaluate alternate masking choices, namely time, frequency, and time-frequency masking, in place of randomly masking audio patches as part of our proposed masking strategy, in table~\ref{tab:asd_altAudMask} and~\ref{tab:sep_altAudMask}. Our model outperforms the versions with these alternatives, showing that random patch masking when combined with channel dropping enables learning more useful features in our setup. This happens possibly because in random patch masking, dropping a full frequency band or time segment is highly improbable thereby allowing our model to extract useful information from the unmasked regions of all frequency bands and time segments of the audio spectrograms.}

\section{Multi-level positional embeddings}\label{sec:multiLevel_PEs}
\begin{table}[!h]
  \centering
  \small
    \setlength{\tabcolsep}{2pt}
    \begin{tabular}{lcc|cc}
    \toprule
                    &   \multicolumn{2}{c|}{TalkNet} &  \multicolumn{2}{c}{SPELL}\\
    Model           & {\textit{EgoCom}} & \multicolumn{1}{c|}{\textit{EasyCom}} & {\textit{EgoCom}} & {\textit{EasyCom}} \\    \midrule
    Ours w/o multi-level PEs       & 59.2 & 70.2 & 60.4 & 65.6 \\
    \textbf{Ours} & \textbf{63.9} & \textbf{71.8} & \textbf{65.6} & \textbf{70.2} \\
    \bottomrule
  \end{tabular}
  \caption{Effect of our multi-level positional embeddings on ASD.
  } \label{tab:asd_multiLevel_PEs}
\end{table}

\begin{table}[!h]
  \centering
    \small
    \setlength{\tabcolsep}{6pt}
    \begin{tabular}{lcc}
    \toprule
    Model           & {SI-SDRi $\uparrow$} & {STFT ($\times 10^{-3}$) $\downarrow$} \\    \midrule
    Ours w/o multi-level PEs & 1.30 & 7.88\\
    \textbf{Ours} & \textbf{2.20} & \textbf{6.51}\\
    \bottomrule
  \end{tabular}
  \caption{Effect of our multi-level positional embeddings on denoising with 0dB noise. 
  } \label{tab:denoising_multiLevel_PEs}
\end{table}

Here, we evaluate the impact of our multi-level positional embeddings by comparing our model with the ablation where positional embeddings are used only at the input level. See table~\ref{tab:asd_multiLevel_PEs} for results on ASD and table~\ref{tab:denoising_multiLevel_PEs} for results on denoising with 0 dB noise. Our model
improves over the ablation on both tasks, showing that using multi-level positional embeddings is crucial for remembering the spatial layout of the tokens at different stages in the model.

\section{Task-specific backbones}\label{sec:taskSpecific_backbones}
\begin{table}[!h]
  \centering
  \small
    \setlength{\tabcolsep}{2pt}
    \begin{tabular}{lcc|cc}
    \toprule
                    &   \multicolumn{2}{c|}{\textit{EgoCom}} &  \multicolumn{2}{c}{\textit{EasyCom}}\\
    Model           & {TalkNet} & \multicolumn{1}{c|}{SPELL} & {TalkNet} & {SPELL} \\    \midrule
    Ours w/o B (from-scratch)       & \multicolumn{2}{c|}{61.1} & \multicolumn{2}{c}{62.0}\\
    Ours w/o B (pretrained)       & \multicolumn{2}{c|}{63.1} & \multicolumn{2}{c}{65.7}\\
    \textbf{Ours} & \textbf{63.9} & \textbf{65.6} & \textbf{71.8} & \textbf{70.2} \\
    \bottomrule
  \end{tabular}
  \caption{Effect of task-specific backbones (denoted using `B') on ASD.
  } \label{tab:asd_taskSpecBackbones}
\end{table}

\begin{table}[!h]
  \centering
    \small
    \setlength{\tabcolsep}{6pt}
    \begin{tabular}{lcc}
    \toprule
    Model           & {SI-SDRi $\uparrow$} & {STFT ($\times 10^{-3}$) $\downarrow$} \\    \midrule
    Ours w/o B (from-scratch)  & 1.02 & 8.99\\
    Ours w/o B (pretrained)  & 2.05 & 7.12\\
    \textbf{Ours} & \textbf{2.20} & \textbf{6.51}\\
    \bottomrule
  \end{tabular}
  \caption{Effect of task-specific backbones (denoted using `B') on denoising with 0dB noise. 
  } \label{tab:denoising_taskSpecBackbones}
\end{table}

Here, we study the impact of using task-specific backbones on our model performance by evaluating two baselines, with the same architecture but without task-specific backbones (Ours w/o B)---one is learned from scratch and another is pretrained. See table~\ref{tab:asd_taskSpecBackbones} for results on ASD and table~\ref{tab:denoising_taskSpecBackbones} for results on denoising with 0 dB noise.Our pretraining scheme leads to better performance than a from-scratch model even w/o B (table~\ref{tab:asd_taskSpecBackbones} and table~\ref{tab:denoising_taskSpecBackbones} top), and we get the best results when we combine our features with task-specific backbones. This shows that while our audio-visual features provide important spatial cues to downstream models, they are not intended to replace the face-specific features used in ASD or the mixed audio features used in denoising.

\subsection{Pretraining model capacity and computational cost}\label{sec:model_capacity}
\begin{table}[t]
  \centering
    \scalebox{0.82}{
    \setlength{\tabcolsep}{4pt}
    \begin{tabular}{lcccc}
    \toprule
                    &   \multicolumn{2}{c}{Model parameter \#} &  \multicolumn{2}{c}{GFLOPs}\\
    Model           & {ASD} & {Denoising} & {ASD} & {Denoising}\\    \midrule  
    2.5D-VS~\citep{gao20192} & 61.2 & 18.2 & 79.5 & 33.2\\
    TLR~\citep{yang2020telling} & 57.5 & 18.1 & 75.9 & 33.7\\
    2.5D-VS~\citep{gao20192}++ & 180.9 & 75.3 & 174.0 & 90.2\\
    AV-MAE~\citep{georgescu2022audiovisual} & 178.6 & 74.1 & 171.6 & 87.5\\
    SSR~\citep{sato2022self-supervised}++ & 180.9 & 75.3 & 174.0 & 90.2\\
    Ours & 180.9 & 75.3 & 174.0 & 90.2\\
    \bottomrule
  \end{tabular}
  }
  \caption{
  {Model parameter count (in millions) and GFLOPs of different pretraining methods.}
  } 
  \label{tab:model_capacity}
\end{table}
{Here, we report the model capacity (parameter count) and GFLOPs of all pretraining methods in Table~\ref{tab:model_capacity}. Note that both the parameter count and GFLOPs of all transformer~\citep{vaswani2017attention}-based methods (2.5DVS~\citep{gao20192}++, AV-MAE~\citep{georgescu2022audiovisual} and SSR~\citep{sato2022self-supervised}++) are comparable to those of our model\footnote{The parameter count and GFLOPs of AV-MAE are a bit lower owing to its modality-inpainting architecture design, where the modality being inpainted is dropped from the input, leading to a slightly smaller model.}, re-emphasizing that our improvements in performance on the downstream tasks are solely attributable to our better model design.}

\subsection{Visual attention maps per audio channel}\label{sec:attn_map_per_ch}
\begin{figure}[!t] 
    \centering
    \includegraphics[width=1\linewidth]{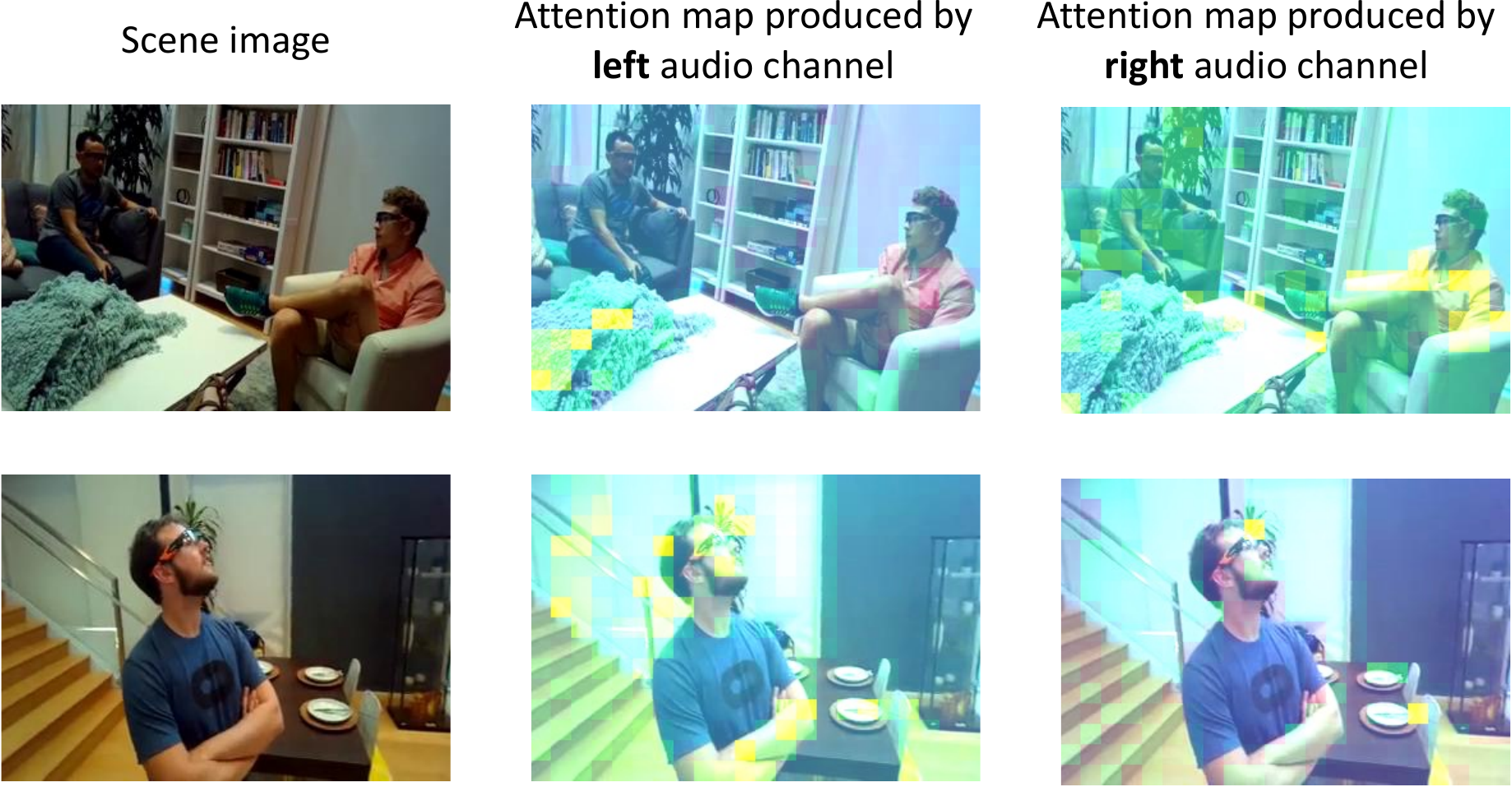}
    \caption{Heat maps for left and right audio channels, similar to Fig. 4 in main.
    Interestingly, our model uses the left channel to focus more on areas to the left of scene image and vice-versa. 
    }
\label{fig:attn_map_lVr}
\end{figure}

{In Fig.~\ref{fig:attn_map_lVr}, we show the attention maps of our model (similar to Fig.~\ref{fig:qual} in main) separately for the left and right channels, on the more challenging EgoCom~\citep{9200754} dataset. We notice that our model uses the left channel to focus more on areas to the left of scene image and vice-versa, indicating that our model can reason about the spatial properties of the scene using both audio and vision.

Further, to better portray the larger trend, we measure the percentage of cases in our test set where the left audio channel attends more towards patches on the left side of the scene image, and the right channel attends more towards patches on the right. This measure comes out to be 62.4\% for the left channel, and 57.2\% for the right channel, showing that our model uses the left channel to focus more on areas to the left of the scene image and vice-versa, across the whole test set.
}

\subsection{Finetuning strategy}\label{sec:finetune_strategy}
\begin{table*}[!t]
\small
  \centering
  \setlength{\tabcolsep}{2pt}
    \begin{tabular}{lrr|rr|rr|rr}
    \toprule
     &  \multicolumn{4}{ c| }{TalkNet~\citep{tao2021someone}} & \multicolumn{4}{ c }{SPELL~\citep{min2022learning}} \\
     &  \multicolumn{2}{ c| }{\textit{EgoCom}} & \multicolumn{2}{ c| }{\textit{EasyCom}} & \multicolumn{2}{ c| }{\textit{EgoCom}} & \multicolumn{2}{ c}{\textit{EasyCom}}\\
    Model & Val & Test & Val & Test & Val & Test & Val & Test\\
    \midrule
    Ours w/ finetuning all audio tokens & 65.3 & \textbf{64.1} & 58.9 & \textbf{71.8} & \textbf{68.4} & 65.5 & 68.7 & 70.1\\
    \textbf{Ours} & \textbf{68.7} & 63.9 & \textbf{60.5} & \textbf{71.8} & \textbf{68.4} & \textbf{65.6} & \textbf{68.9} & \textbf{70.2}\\
    \bottomrule
  \end{tabular}
  \caption{{ASD with our model when all tokens are used in downstream training.}}
  \label{tab:asd_altFntn}
\end{table*}

\begin{table*}[!t]
\small
  \centering
  \setlength{\tabcolsep}{2pt}
    \begin{tabular}{lrr|rr|rr}
    \toprule
     &  \multicolumn{2}{ c| }{0 dB} & \multicolumn{2}{ c| }{2.5 dB} & \multicolumn{2}{ c}{5 dB}\\
    Model & SI-SDRi $\uparrow$ & STFT $\downarrow$ & SI-SDRi $\uparrow$ & STFT $\downarrow$ & SI-SDRi $\uparrow$ & STFT $\downarrow$\\
    \midrule
    Ours w/ finetuning all audio tokens & 2.18 & \textbf{6.47} & 2.50 & 4.49 & \textbf{2.58} & \textbf{2.52}\\
    \textbf{Ours} & \textbf{2.20}& 6.51 & \textbf{2.72} &\textbf{4.22} & 2.46 & 2.70\\ 
    \bottomrule
  \end{tabular}
  \caption{{Denoising with our model when all audio tokens are used in downstream training. All STFT distance measures use base $10^{-3}$.}}
  \label{tab:sep_altFntn}
\end{table*}

{We mask audio tokens during finetuning primarily to reduce the computation overhead. Since our pretraining also involves token masking, our model can learn strong audio-visual features even when all audio tokens are not available during finetuning. However, to quantatively evaluate the effect of our finetuning strategy, we also finetune our model with all audio tokens and report the results in table~\ref{tab:asd_altFntn} for ASD and table~\ref{tab:sep_altFntn} for denoising. We don’t see a significant change in performance when using all the tokens compared to masking when finetuning, but using all tokens is 1.7 times slower on average.}

\subsection{Model parameter initialization}\label{sec:param_init}
To evaluate the effect of random parameter initialization on our model, we train our model on both tasks and datasets with 3 different random seeds. Across all runs, our standard errors are less than 0.01 on all metrics, showing that our model is robust to different random parameter initializations, and the improvements in performance are significantly larger than these small variations from randomness.

\subsection{Dataset details}\label{sec:supp_data}
As discussed in main (Sec.~\ref{sec:experiments}), we use two public datasets containing egocentric videos with binaural audio, EgoCom~\citep{9200754} and EasyCom~\citep{donley2021easycom}, for our experiments. For EgoCom, we follow the authors and split the data into train/val/test comprising 30.3/2.4/5.8 hours of data. For EasyCom, we randomly generate train/val/test splits with 4.5/0.38/0.39 hours of data, such that there is no overlap in conversation participants between any two splits. Next, we extract 1 second long clips from both datasets, where the video and binaural audio are sampled at 5 frames per second (fps) and 16 kHz, respectively. The frame resolution is $240 \times 352$ for EgoCom, and $198 \times 352$ for EasyCom. Furthermore, we choose audio channel 5 and 6 (corresponding to the in-ear microphones) as our binaural audio channels for EasyCom.

\subsection{Model architecture and training details}\label{sec:modelArc_n_training}
In addition to the provided details in {Sec.~\ref{sec:training},~\ref{sec:main_asd} and~\ref{sec:main_sad} in} main,
we provide here extra model architecture and training details for both pretraining and finetuning on downstream tasks, for reproducibility. We perform all training using 8 NVIDIA Tesla V100 SXM2 GPUs. We will release all code and data.

\subsubsection{Pretraining} 
We described our model architecture and pretraining details in Sec.~\ref{sec:training} in main. Here, we provide additional details about our model's input preparation, architecture, parameter initialization, and training . 

\paragraph{Input preparation.}
We sample the video clips at their original resolution, normalize them using the per-color means and standard deviations computed using ImageNet~\citep{he2016deep}, and generate a total of 330 and 286 visual tokens for EgoCom and EasyCom, respectively, by splitting the clips into non-overalapping tubelets containing a sequence of 5 patches, where each patch is $16 \times 16$ in size (L{193-5} in main). We represent the binaural audio as two-channel Kaldi-compliant~\citep{Povey_ASRU2011} spectrograms with $98$ temporal windows and $128$ Mel-frequency bins, which we compute by using the binaural audio normalized to $[-1, 1]$, a window length of 25 ms and a hop length of 10 ms. We normalize the spectrograms by computing the mean and standard deviation of the Mel-spectrograms generated from all audio clips in each dataset.  We next generate  392 audio tokens per spectrogram channel by splitting it into non-overlapping patches of size $2 \times 16$. 

\paragraph{Architecture.} All hidden layers in each transformer block~\citep{feichtenhofer2022masked} emit features that are four times as long as the embedding size for the block. We always use LayerNorm~\citep{ba2016layer} after every output of a transformer block unless it's a direct input to another transformer block. 

\paragraph{Parameter initialization.} We use Xavier~\citep{Glorot2010UnderstandingTD} uniform initialization for all network parameters. For the LayerNorm~\citep{ba2016layer} layers, we initialize their weights to 1 and biases to 0. We use a truncated normal distribution with a standard deviation of 0.02 and a sampling range of $[-2, 2]$ to initialize the learnable modality and channel embedding tokens, and initialize the mask tokens from a normal distribution with a standard deviation of 0.02.

\paragraph{Training.} We set the batch size to 104 and weight decay to $10^{-5}$ during pretraining.

\subsubsection{Active speaker detection}

\begin{figure*}[t] 
    \centering
    \includegraphics[width=0.75\linewidth]{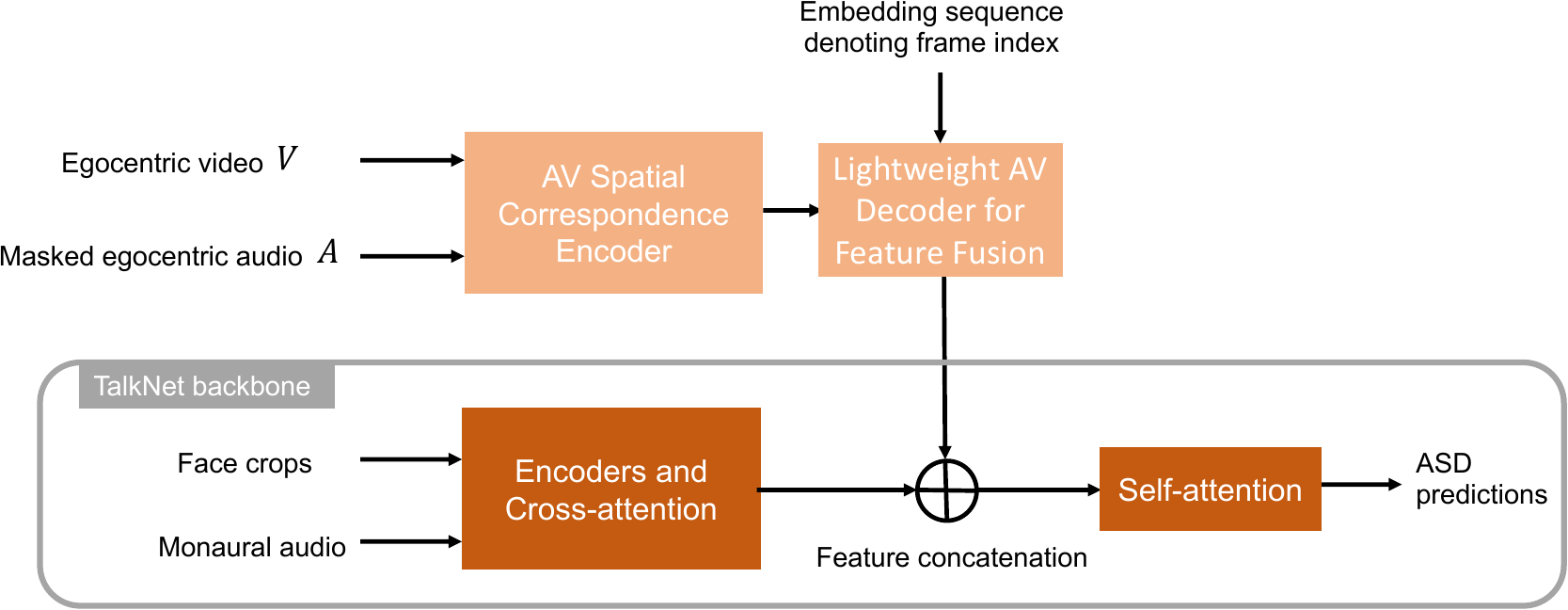}
    \caption{Method to fuse our pretrained features with TalkNet~\citep{tao2021someone} for ASD.}
\label{fig:feat_fusion_talknet}
\end{figure*}

\begin{figure*}[t] 
    \centering
    \includegraphics[width=0.75\linewidth]{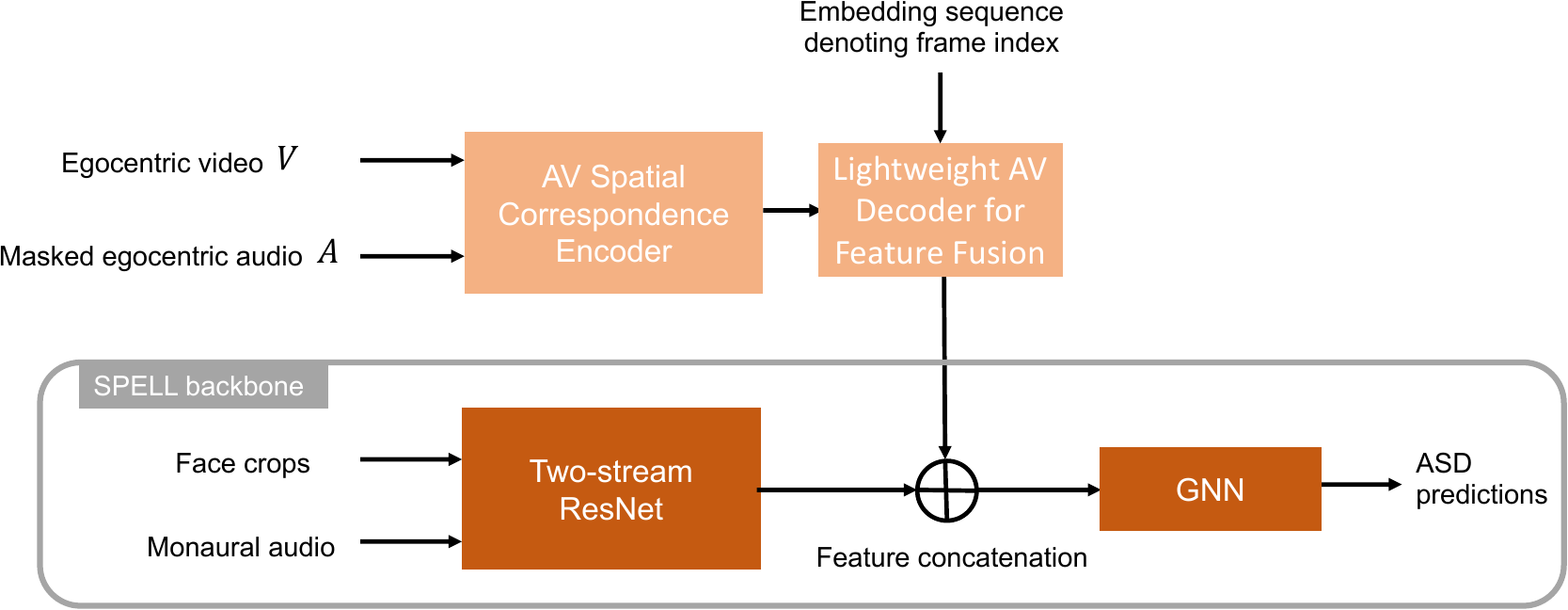}
    \caption{Method to fuse our pretrained features with SPELL~\citep{min2022learning} for ASD.}
\label{fig:feat_fusion_spell}
\end{figure*}

\begin{figure*}[t] 
    \centering
    \includegraphics[width=0.75\linewidth]{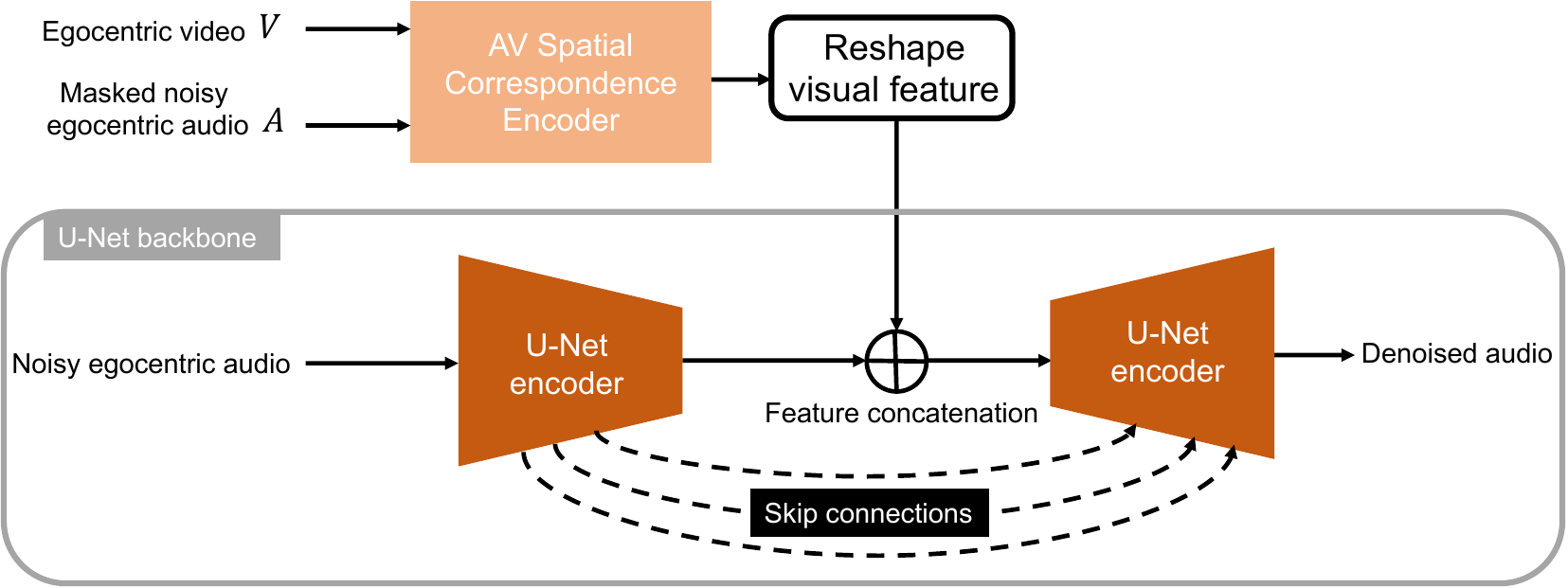}
    \caption{Method to fuse our pretrained features with U-Net~\citep{yang2020telling} for spatial audio denoising.}
\label{fig:feat_fusion_unet}
\end{figure*}

In Sec.~\ref{sec:main_asd} in main, we outlined our feature fusion method for active speaker detection (ASD). Here, we provide additional architectural details for feature fusion, and also describe our finetuning process.

\paragraph{Pretrained feature fusion.}
Fig.~\ref{fig:feat_fusion_talknet} and~\ref{fig:feat_fusion_spell} show our feature fusion methods for TalkNet~\citep{tao2021someone} and SPELL~\citep{min2022learning} ASD backbones, respectively. The single-layer transformer decoder (Sec.~\ref{sec:main_asd} in main), which we use for fusing our pretrained features with the backbones (Sec.~\ref{sec:main_asd} in main), generates 128 and 512 dimensional embeddings for TalkNet
and SPELL,
respectively. Since SPELL doesn't train any audio-visual features when training its graph neural network (GNN), we first pretrain the the transformer decoder for SPELL by using it with the TalkNet backbone. Towards that goal, we feed the decoder features to a single linear layer that maps the 512 dimensional features to 128 dimensional features, and is followed by GELU~\citep{Hendrycks2016GaussianEL} activations and LayerNorm~\citep{ba2016layer}, before fusing the 128 dimensional features with the TalkNet backbone. After pretraining, we append the 512 dimensional outputs of the decoder with the outputs of the two-stream audio-visual encoder (L{405-8} in main) for training the GNN in SPELL.

\paragraph{Training.}
For TalkNet, we train using Adam~\citep{DBLP:journals/corr/KingmaB14} for 25 epochs optimizer with an initial learning rate (LR) of $10 ^ {-4}$ for the backbone and $10 ^ {-5}$ for the pretrained components, both of which we decay using a step LR scheduler by a factor of 0.95 after every epoch. We set the batch size to 400.

For SPELL, we first train the two-stream audio-visual encoder for feature extraction for 100 epochs using the cross entropy loss and Adam~\citep{DBLP:journals/corr/KingmaB14} with an initial learning rate of $5 \times 10^{-4}$, which we decay by $0.1$ after every 40 epochs. We set the batch size to 320. For training the GNN of SPELL, we train for 70 epochs by using a batch size of 320 again and learning rate of $10^{-3}$, while setting all other hyperparameters per the original paper.

\subsubsection{Spatial audio denoising}
\paragraph{Backbone architecture.}

Following~\citep{yang2020telling}, our U-Net backbone for spatial audio denoising (Sec.~\ref{sec:main_sad} in main) is an audio-visual model comprising an audio encoder, a visual encoder, and a decoder for predicting an estimate of the target audio. The audio encoder takes the log magnitude spectrogram of the mixed binaural audio as input, and uses a stack of 5 convolutional (conv.) layers to produce a multi-channel 2D audio feature map. Each conv. layer has a kernel size of 4, padding of 1, and stride of 2, and is followed by leaky ReLU~\citep{nair2010rectified} activations with a slope of 0.2 for negative inputs, and batch normalization~\citep{pmlr-v37-ioffe15}. The conv. layers have 64, 128, 256, 512 and 512 output channels, respectively. The visual encoder has a ResNet-18~\citep{he2016deep} architecture that outputs a multi-channel 2D visual feature map without feeding it to the average pooling or any subsequent layers. We push the ResNet outputs through another conv. layer to match its height and width with the audio features. The conv layer has a kernel size of (1, 4), a padding of (0, 0) for EgoCom~\citep{9200754} and (1, 0) for EasyCom~\citep{donley2021easycom}, and 128 output channels. Further, we remove the last feature column from the output of the conv. layer for all channels for EasyCom. We concatenate the per-frame features along the channel dimension and generate the visual features. We then concatenate the visual features with the audio features channel-wise, and feed the concatenated features to the audio decoder, which predicts an estimate of the ratio mask~\citep{gao20192, yang2020telling} for the target audio magnitude spectrogram. The audio decoder first uses a stack of 5 transpose convolutional (conv.) layers, which are connected to the corresponding encoder layers through skip connections. The transpose conv. layers have a kernel size of 4, stride of 2, and a padding of (1, 1), except for the last layer, which has a padding of (2, 1). The transpose conv. layers have 1152, 1024, 512, 256 and 128 output channels, respectively. Next, the audio decoder feeds the output of the transpose conv. layers to a conv. layer with 2 input and output channels, and a kernel size of (2, 1) to emit the predicted ratio mask.

\paragraph{Input preparation.}
To transform the audio waveforms into magnitude spectrograms, we first normalize them to [-1, 1] and then compute the short-time Fourier transform with a window length of 128, hop length of 64, and 512 frequency bins.

\paragraph{Pretrained feature fusion.}
Fig.~\ref{fig:feat_fusion_unet} shows our feature fusion method for spatial audio denoising. We reshape the visual features from the outputs of our audio-visual encoder $\mathcal{E}^{AV}$ to form multi-channel 2D visual feature maps (Sec.~\ref{sec:main_sad} in main), such that the 2D raster order of the features matches that of the tubelets in the video clip, and feed the reshaped features to a convolutional (conv.) layer with a kernel size of (3, 4), stride of (2, 3), padding of (1, 2) and (2, 2) for EgoCom~\citep{9200754} and EasyCom~\citep{donley2021easycom}, respectively, and 128 input and 784 output channels. We similary reshape the audio features, and feed them to another conv. layer with a kernel size of (1, 7), padding of 0, stride of (1, 6), and 128 input and 256 output channels. Both conv. layers are followed by leaky ReLU activations with a slope of 0.2 for the negative values, and batch normalization. Next, we concatenate the visual and audio features along the channel dimension, and further concatenate them with the audio encoder outputs channel-wise (Sec.~\ref{sec:main_sad} in main).

\paragraph{Training.}
We train using Adam~\citep{DBLP:journals/corr/KingmaB14} for 200 epochs optimizer with an learning rate (LR) of $5 \times 10 ^ {-4}$. We set the batch size to 80.


\end{document}